\pdfoutput=1

\documentclass[11pt]{article}

\usepackage[final]{acl}

\usepackage{times}
\usepackage{latexsym}

\usepackage[T1]{fontenc}

\usepackage[utf8]{inputenc}

\usepackage{microtype}

\usepackage{inconsolata}

\usepackage{graphicx}
\usepackage{times}
\usepackage{latexsym}
\usepackage{times}
\usepackage{latexsym}
\usepackage{booktabs}
\usepackage{boxedminipage}
\usepackage{amssymb}
\usepackage{amsmath}
\usepackage{todonotes} 
\usepackage{graphicx}

\usepackage{hyperref}
\usepackage{inconsolata}
\usepackage{pifont}

\usepackage{tabu}
\usepackage{fontawesome}
\usepackage{amsmath}
\usepackage{amsfonts}
\usepackage{amssymb}
\usepackage{enumitem}
\usepackage{listings}
\usepackage{xstring}
\usepackage{graphicx}
\usepackage{pbox}
\usepackage{subcaption}
\usepackage{epstopdf}
\usepackage{xstring}
\usepackage{multirow}

\usepackage{soul}
\usepackage{color}
\usepackage{graphicx}
\usepackage{multirow}
\usepackage{comment}
\usepackage{todonotes}
\usepackage{listings} 
\usepackage{graphicx}
\usepackage{subcaption}
\usepackage{verbatim}

\lstdefinestyle{smalllisting}{
    basicstyle=\small\ttfamily
}

\usepackage{xcolor} 
\definecolor{lightblue}{rgb}{.50,.90,0.51}
\definecolor{tri}{rgb}{.25,.88,.82}
\definecolor{lilac}{rgb}{0.85,0.64,0.85}
\definecolor{atomictangerine}{rgb}{1.0, 0.6, 0.4}

\newcommand{\memex}{\emph{MemeXplain}}

\title{MemeIntel: Explainable Detection of Propagandistic and Hateful Memes}

\author{Mohamed Bayan Kmainasi$^1$\thanks{~The contribution was made while the author was a contributor at the Qatar Computing Research Institute.}\textsuperscript{$\dagger$}, Abul Hasnat$^2$$^,$$^3$\thanks{~Equal contribution.}, Md Arid Hasan$^1$$^*$,\\
{\bf Ali Ezzat Shahroor$^5$, Firoj Alam$^5$\textsuperscript{$\dagger$}\thanks{~Corresponding author.}}\\
  $^1$Qatar Computing Research Institute, Qatar, 
  $^2$Blackbird.AI, USA, 
  $^3$APAVI.AI, France \\
\texttt{mk2314890@qu.edu.qa, mhasnat@gmail.com,}\\
\texttt{arid.hasan@unb.ca, \{fialam, alsh34060\}@hbku.edu.qa}
\\}

\begin{document}
\maketitle
\begin{abstract}
The proliferation of multimodal content on social media presents significant challenges in understanding and moderating complex, context-dependent issues such as misinformation, hate speech, and propaganda. While efforts have been made to develop resources and propose new methods for automatic detection, limited attention has been given to jointly modeling label detection and the generation of explanation-based rationales, which often leads to degraded classification performance when trained simultaneously. To address this challenge, we introduce \memex{}, an explanation-enhanced dataset for \textbf{propagandistic memes in Arabic and hateful memes in English}, making it the \textit{first} large-scale resource for these tasks. To solve these tasks, we propose a \textbf{multi-stage optimization approach} and train \textbf{Vision-Language Models (VLMs)}. Our results show that this strategy significantly improves both \textbf{label detection} and \textbf{explanation generation} quality over the base model, outperforming the current state-of-the-art with an \textbf{absolute improvement of $\sim{1.4}$\% (Acc)} on ArMeme and $\sim{2.2}$\% (Acc) on Hateful Memes.
For reproducibility and future research, we aim to make the \memex{} dataset and scripts publicly available.\footnote{\url{https://github.com/MohamedBayan/MemeIntel}}
\end{abstract}

\section{Introduction}
\label{sec:introduction}

Despite the rapid growth of multimodal content—integrating images, text, and sometimes video—the automated detection of harmful and false information on online news and social media platforms has become increasingly critical. In particular, identifying propaganda and hate in memes is essential for combating misinformation and minimizing online harm. While most research has focused on textual analysis, multimodal approaches have received comparatively less attention. In propaganda detection, text-based methods have evolved from monolingual to multilingual setups \cite{piskorski-etal-2023-multilingual,hasanain-etal-2023-araieval}, from binary to 
fine-grained span-level tasks \cite{BARRONCEDENO20191849,Habernal.et.al.2017.EMNLP,Habernal2018b,da-san-martino-etal-2019-fine}. Hate speech detection has similarly progressed from text-based to multimodal approaches that integrate both textual and visual elements \cite{kiela2020hateful,alam2022survey,ijcai2022p781}. 

\begin{figure}[]
    \centering
    \includegraphics[scale=0.23]{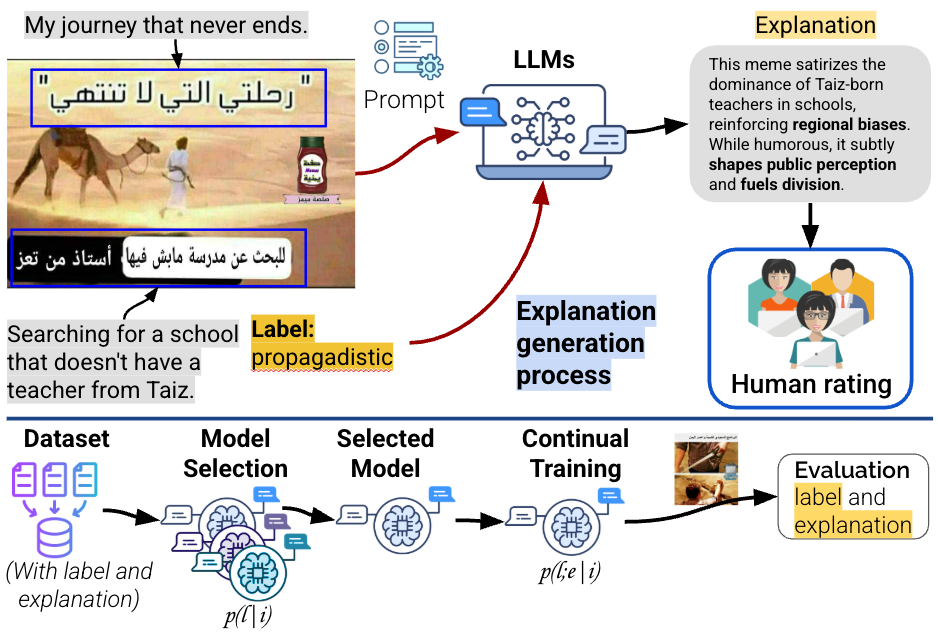}
    \vspace{-0.2cm}
    \caption{Experimental steps for explanation generation and training.}    
    \label{fig:vllm_exp_meme}
    \vspace{-0.2cm}
\end{figure}


The emergence of LLMs has demonstrated significant capabilities across various disciplines. To capture implicit hate and propaganda, LLMs utilize different techniques \cite{cao-etal-2022-prompting, kumar2022hate}, such as prompt-based methods, contrastive learning, and cross-modal attention. Consequently, efforts have been made to leverage VLMs \cite{zhang2024vision} to classify the harmful and propagandistic memes \cite{Rui2023}.
Despite progress, detecting implicit hate, especially with sarcasm or irony, remains challenging. Propagandistic memes add complexity through emotional appeals, humor, cultural cues, and manipulative language. These challenges are especially pronounced for English-centric multilingual models, which often struggle to capture nuances in non-English content due to cultural differences and linguistic variation. Specific challenges for Arabic often include cultural and political references that do not translate well into Western contexts, which makes it difficult for general-purpose models to detect accurately. For instance, memes about local conflicts or those using religious symbols often rely on subtle cues that are deeply rooted in cultural context. Arabic-focused datasets have made it clear that understanding these nuances is essential for accurate detection and meaningful explanations \cite{alam-etal-2024-armeme}. 
To address these nuances, it is crucial for a system to provide not only accurate predictions but also clear, relatable explanations that align with the meme’s visual context and aid user understanding \cite{hee2023decoding, yang2023hare, huang_chain_2023, sun_text_2023}.


An explanation-based approach offers numerous advantages and improves performance in various tasks \cite{li2022explanations, magister2022teaching, nandi2024safe, kumari2024m3hop}. Although most studies have focused on textual content \cite{li2022explanations, magister2022teaching}, some recent approaches \cite{nandi2024safe, kumari2024m3hop} have applied explainability to images. However, these methods rely on QA-based explanations that lack naturalness, use multiple inference with custom models, thus increasing computational complexity, and employ explanations only during training rather than as an inference output. To address these limitations, we propose a simplified method that delivers strong performance in meme classification and explanation generation across two datasets.

We summarize our main contributions below: 
\begin{itemize}[noitemsep,topsep=0pt,labelsep=.5em]
    \item We developed an explanation-enhanced datasets, \memex{}, using a rapid and low-cost annotation procedure;
    \item We investigated state-of-the-art VLMs to identify an appropriate model for meme classification and explanation generation;
    \item We proposed an efficient \textit{multi-stage optimization procedure} that mitigates gradient conflicts and avoids catastrophic forgetting by formulating the optimization problem from the perspective of \emph{domain adaptation} and \emph{task‑incremental learning}. 
    \item We achieved state-of-the-art performance on two types of datasets related to propaganda and hateful content detection.
\end{itemize}
Our findings are as follows:
\textit{(a)} A higher human evaluation score suggests that explanations from stronger models (e.g., GPT-4o) are reliable and can serve as gold-standard explanations for training smaller models; \textit{(b)} Task-specific fine-tuning improves performance over the base model; and \textit{(c)} Our multi-stage optimization improves label detection and explanation quality by reducing gradient conflicts and avoiding catastrophic forgetting, achieving state-of-the-art performance.
%
Overall, our work is the \textbf{\textit{first}} to enhance VLMs for simultaneous propaganda and hateful content detection while providing natural reasoning to end users.




\section{Related Work}
\label{sec:related_work}

The widespread use of social networks has become a major channel for spreading misinformation, propaganda, and harmful content. Significant research efforts have been directed toward addressing these challenges, particularly in multimodal disinformation detection~\cite{alam2022survey}, harmful memes~\cite{ijcai2022p781}, and propagandistic content~\cite{ACL2021:propaganda:memes}. However, most studies have focused on detection, while less attention has been given to generating natural explanations/reasons behind the predicted labels.

\noindent
\paragraph{Multimodal Propagandistic Content.}
Following the previous research for propaganda detection using textual content~\cite{da-san-martino-etal-2019-fine}, \citet{dimitrov-etal-2021-semeval} introduced SemEval-2021 Task 6 focusing on persuasion techniques detection in both textual and visual memes. Subsequently, the focus has extended to the detection of multilingual and multimodal propagandistic memes \cite{dimitrov2024semeval}. 
Similar multimodal work on Arabic involves the development of datasets and shared task for propaganda detection \cite{alam-etal-2024-armeme, araieval:arabicnlp2024-overview}. For the detection problem, typical approaches include a fusion of textual and visual embedding and a classification head on top them~\cite{araieval:arabicnlp2024-overview,shah-etal-2024-mememind}, graph attention network based approach
for multimodal
objects \citet{chen2024multimodal}.

\noindent
\paragraph{Multimodal Hate speech.}
Similarly, there has been growing interest in detecting multimodal hate speech \cite{kiela2020hateful, velioglu2020detecting, hee2022explaining}. Due to the lack of resources, \citet{kiela2020hateful} developed a large-scale dataset for multimodal hate identification. This study advanced research in this area and emphasized the importance of integrating textual and visual features for effective detection.
To further progress in this field, efforts have been made to develop resources for multiple languages, including Arabic \cite{alam2024propaganda}, Bangla \cite{hossain-etal-2022-mute}, and English \cite{hee2023decoding}. A more detailed summary 
can be found in \citet{ijcai2022p781}, which also highlights key challenges and outlines future research directions.

\noindent
\paragraph{Training with Explanations.}
Integrating reasoning or explainable capabilities to enhance LLM/VLM performance has been shown to be highly beneficial for various tasks across multiple domains \cite{plaat2024reasoning}. This approach has also proven effective for knowledge distillation and model compression \cite{li2022explanations, magister2022teaching}, where explanations generated by large LLMs improve the performance and capabilities of smaller LLMs. 
%
For example, in the hate-speech detection task, Hare \cite{yang2023hare} employs Chain-of-Thought (CoT) reasoning, while \citet{huang_chain_2023} utilizes Chain of Explanation (CoE). Their aim is to improve the effectiveness of LLM-based sentiment classifiers by leveraging reasoning capabilities. \citet{sun_text_2023} proposed Clue and Reasoning Prompting (CARP), which uses keywords and reasoning to guide explanation.
%
%
\noindent
\paragraph{CoT-Based Approaches.}
CoT is a widely recognized prompting technique that generates a chain of reasoning to derive answers. 
%
\citet{kumari2024m3hop} proposed a CoT-based framework for meme analysis using scene graphs to capture text- and image-based entity-object relationships. Their three-step prompting strategy guides the LLM to identify Emotion, Target, and Context for effective meme interpretation. SAFE-MEME \cite{nandi2024safe} introduced two multimodal datasets and a CoT-based reasoning framework for meme-based hate speech detection, using Q\&A-style prompts and hierarchical labels. However, their method was not evaluated on the Hateful Memes dataset, limiting direct comparison. 
\noindent
\paragraph{Comparison with Prior Work.}
A key limitation of CoT-based methods is their reliance on multi-step reasoning, requiring multiple VLM inferences. In contrast, our approach: \textit{(a)} avoids complex CoT steps, improving efficiency and reducing cost; and \textit{(b)} generates explanations alongside classifications to enhance transparency and reliability.
%
%
\citet{hee2023decoding} constructed a dataset providing explanations for hateful memes. However, unlike us, they focused solely on evaluating explanation generation and did not perform classification tasks. 
Despite the availability of their data, we do not use it due to the lack of \textit{naturalness}. In particular, their explanations do not fully account for image content or image-centric contextual perspectives.

%
\section{Dataset}
\label{sec:dataset}
\subsection{ArMeme}
The ArMeme dataset 
comprises approximately $\sim$6k manually annotated Arabic memes collected from various social media platforms
\cite{alam-etal-2024-armeme}. This dataset has been 
manually annotated with four labels such as \textit{Not propaganda}, \textit{Propaganda}, \textit{Not-meme} and \textit{Other}. Table \ref{tab:data_split_stat} provides the distribution of the data splits. The memes with ``Not propaganda'' category covers over half of the dataset ($\sim$66\%), followed by ``Propaganda'' and the distribution of ``Not-meme`` and ``Other`` classes are significantly smaller. This distribution highlights a substantial class imbalance, particularly between ``Not propaganda'' and the other categories. 


\begin{table}[h]
\centering
\setlength{\tabcolsep}{3pt} 
\scalebox{0.95}{%
\begin{tabular}{@{}lrrrr@{}}
\toprule
\multicolumn{1}{c}{\textbf{Class label}} & \multicolumn{1}{c}{\textbf{Train}} & \multicolumn{1}{c}{\textbf{Dev}} & \multicolumn{1}{c}{\textbf{Test}} & \multicolumn{1}{c}{\textbf{Total}} \\ \midrule
Not propaganda & 2,634 & 384 & 746 & 3,764 \\
Propaganda & 972 & 141 & 275 & 1,388 \\
Not-meme & 199 & 30 & 57 & 286 \\
Other & 202 & 29 & 56 & 287 \\ \midrule
\textbf{Total} & \textbf{4,007} & \textbf{584} & \textbf{1,134} & \textbf{5,725} \\ \bottomrule
\end{tabular}
}
\vspace{-0.2cm}
\caption{Data splits for ArMeme datasets.}
\label{tab:data_split_stat}
\vspace{-0.3cm}
\end{table}

\subsection{Hateful Meme}
The Hateful Memes dataset~\cite{kiela2020hateful}, is a benchmark designed to evaluate multimodal hate speech detection. It consists of $\sim$12k memes, combining both text and images, carefully curated to ensure that effective classification requires an understanding of both modalities. The dataset was created using a mix of synthetically generated memes and real-world examples, sourced from social media, while ensuring a balanced distribution of hateful and non-hateful content. 
In Table \ref{tab:data_stat_hateful_meme}, we report the distribution of hateful meme dataset used for this study. Note that hateful meme dataset consists of two other splits (dev-seen and test-seen), here, we used unseen versions.


\begin{table}[h]
\centering
\setlength{\tabcolsep}{3pt}
\scalebox{0.9}{%
\begin{tabular}{@{}lrrrr@{}}
\toprule
\multicolumn{1}{c}{\textbf{Class Label}} &
\multicolumn{1}{c}{\textbf{Train}} &
\multicolumn{1}{c}{\textbf{Dev-unseen}} &
\multicolumn{1}{c}{\textbf{Test-unseen}} &
\multicolumn{1}{c}{\textbf{Total}} \\ \midrule
Not Hateful & 5,481 & 340 & 1,250 & 7,071 \\
Hateful     & 3,019 & 200 &   750 & 3,969 \\ \midrule
\textbf{Total} & \textbf{8,500} & \textbf{540} & \textbf{2,000} & \textbf{11,040} \\ \bottomrule
\end{tabular}
}
\vspace{-0.2cm}
\caption{Distribution of hateful-meme dataset.}
\label{tab:data_stat_hateful_meme}
\vspace{-0.35cm}
\end{table}

\section{\memex{}: Explanation Generation}
\label{sec:experiments_expl}
%
The outcomes of an automatic system become more reliable for users if it provides decisions with adequate and interpretable natural explanations, which help users better understand the underlying reason behind the system’s decision \cite{hee2023decoding, yang2023hare, huang_chain_2023, sun_text_2023}. Technically, this approach provides numerous advantages in terms of knowledge distillation, model compression, and enhancing the performance of target tasks in different domains \cite{li2022explanations, magister2022teaching, nandi2024safe, kumari2024m3hop}. This motivates us to adopt the explanation-based approach in our research. However, we also aim to improve its efficiency, particularly with respect to dataset generation, model training, and system inference procedures.

In this research, we generate explanations for two different stages: 
(a) during existing dataset enhancement, which leverages an expert 
VLM (such as GPT-4o) to generate high-quality explanations and 
(b) during training/inference with a smaller VLM (such as Llama-3.2 11b). 
Mathematically, these two stages can be described by the functions
$f(i, l)=e$ and $g(i)=(l,e)$, where \( e \) denotes the explanation, 
\( l \) is the label, and \( i \) is the input image or meme (can also be associated text). 
Specifically, \( f(i, l) \) is defined as the function that generates an explanation \( e \) for dataset enhancement, given an image \( i \) and its corresponding label  \( l \) as input; and \( g(i) \) is defined as the function that generates a label-explanation pair \( (l, e) \) during training and inference, given only the image content \( i \) as input. 
In the data preparation phase, we use a VLM to generate explanations conditioned on both the input and the label using the function \( f \). Next, in the training phase, the model learns to jointly predict the label and explanation from the input alone using the function \( g \).

%
%

This research enhances two existing datasets
with explanations, see Section \ref{sec:dataset} and Table \ref{tab:explation_stat} for the details and statistics. For the explanation generation task, it first uses a VLM for \( f(i, l) \) and then involves human experts, which significantly accelerates high-quality explanation generation and lowers the overall cost and time. The following subsections provide step-by-step details.

\begin{table}[]
\centering
\setlength{\tabcolsep}{4pt} 
\scalebox{0.9}{%
\begin{tabular}{@{}lrr|rrrr@{}}
\toprule
\textbf{Data} & \multicolumn{1}{c}{\textbf{\begin{tabular}[c]{@{}c@{}}Total \\ Words\end{tabular}}} & \multicolumn{1}{c}{\textbf{\begin{tabular}[c]{@{}c@{}}Avg. \\ Words\end{tabular}}} & \multicolumn{2}{|c}{\textbf{\begin{tabular}[c]{@{}c@{}}Total Expl. \\ Words\end{tabular}}} & \multicolumn{2}{c}{\textbf{\begin{tabular}[c]{@{}c@{}}Avg. \\ Expl. \\ Words\end{tabular}}} \\ \cline{4-7}
\multicolumn{3}{c|}{} & \multicolumn{1}{c}{\textbf{Ar}} & \multicolumn{1}{c}{\textbf{En}} & \multicolumn{1}{c}{\textbf{Ar}} & \multicolumn{1}{c}{\textbf{En}} \\ \midrule
\multicolumn{7}{c}{\textbf{ArMeme}} \\ \midrule
Train & 58,688 & 15 & 280,341 & 375,843 & 70 & 94 \\
Dev & 8,583 & 15 & 40,756 & 55,336 & 70 & 95 \\
Test & 16,653 & 15 & 79,360 & 105,476 & 70 & 93 \\ \midrule
\textbf{Total} & \textbf{83,924} & \textbf{15} & \textbf{400,457} & \textbf{536,655} & \textbf{70} & \textbf{94}  \\ \midrule
\multicolumn{7}{c}{\textbf{Hateful Meme}} \\ \midrule
Train & 99,812 & 12 & -- & 740,624 & -- & 87 \\
Dev & 4,904 &	9	& -- & 43,956 & -- & 81 \\
Test & 18,079 & 9 & -- & 173,982 & -- & 87 \\ \midrule
\textbf{Total} & \textbf{122,795} & \textbf{10} & -- & \textbf{958,562} & -- & \textbf{85} \\ \bottomrule
\end{tabular}
}
\vspace{-0.2cm}
\caption{Descriptive statistics of the dataset. \textit{Total Words} and \textit{Avg.} refer to the total and average number of words in the text. The last two columns represent the corresponding values for the explanations.
}
\label{tab:explation_stat}
\vspace{-0.35cm}
\end{table}

\subsection{VLMs for Explanation Generation}
Figure \ref{fig:vllm_exp_meme} illustrates an example of an Arabic meme 
along with its explanation-generation process using a VLM. We leverage GPT-4o (version 2024-11-20) for automated explanation generation. The choice of this model is motivated by prior studies \citet{wang_evaluating_2023}, which show that advanced GPT models can produce fluent, informative, persuasive, and logically sound explanations when properly prompted. In Listing~\ref{lst:prompt_explanation_generation_armeme_ar_expl}, we present the \textit{prompts} used for generating explanations for \textbf{ArMeme} and \textbf{Hateful Memes}. To refine the prompt, we iteratively tested several memes in both English and Arabic, selecting the one that produced the most reasonable explanations.

For Arabic memes, we generate two sets of explanations—one \textit{in English} and one \textit{in Arabic}. The motivation behind this approach is to assess the multilingual capability and quality of smaller VLMs, such as Llama-3.2 11b, in generating explanations and labels in both languages. 
\color{black}


\paragraph{Size of the Explanation}

Determining the optimal length for the explanations is important to balance informativeness and cognitive load~\cite{herm2023impact}. \citet{shen2022shortest} explored the relationship between the length of the explanation and human understanding, finding that the shortest rationales are often ineffective. Recently, \citet{wang_evaluating_2023} also studied the effect of explanation size and found that human evaluators are less willing to read longer explanations. To achieve an optimal balance, we iteratively tested various explanation lengths and ultimately set a limit of 100 words.

\paragraph{Model and its Parameters}
To utilize GPT-4o~\cite{openai2023gpt4}, we accessed the OpenAI API via Azure services. Although recently released o1 models have shown promising directions for complex reasoning, they were not accessible to us. For explanation generation, we used zero-shot inference. To ensure reproducibility, we set the temperature value to zero.


\subsection{Manual Annotation of Explanation}
\label{ssec:human_eval}
Given that our idea is to use the generated explanation as gold data for further training and evaluation, therefore, we intended to go through human evaluation process. Following the prior studies \cite{wang_evaluating_2023,huang_chain_2023,agarwal_faithfulness_2024} we adopted four metrics discussed below. For each metric we use 5-point Likert scale. 

\noindent
\textbf{Informativeness.} Measures the extent to which the explanation provides relevant and meaningful information for understanding the reasoning behind the label. A highly informative explanation offers detailed insights that directly contribute to the justification, while a low-informative explanation may be vague, incomplete, or lacking key details.

\noindent
\textbf{Clarity.} Assesses how clearly the explanation conveys its meaning. A clear explanation is well-structured, concise, and easy to understand without requiring additional effort. It should be free from ambiguity, overly complex language, or poor phrasing that might hinder comprehension.

\noindent
\textbf{Plausibility.} Refers to the extent to which an explanation logically supports the assigned label and appears reasonable given the meme's content. A plausible explanation should be coherent, factually consistent, and align with the expected reasoning behind the label.

\noindent
\textbf{Faithfulness.} Measures how accurately an explanation reflects the reasoning behind the assigned label. A faithful explanation correctly represents the key factors and logical steps that justify the label, without adding misleading or unrelated details. 

\noindent\textbf{Annotation Setup.}
For manual annotation, we first prepared an annotation guideline and a platform (see Appendix \ref{sec:app_annotation_platform} and \ref{sec:app_annotation_guideline}, respectively) for the annotators. 
%
%
For the Arabic memes, we recruited annotators who are native Arabic speakers and fluent in English, all holding at least a bachelor's degree. Because of their fluency, they also handled the hateful meme. We provided necessary training and consultation, and all had prior experience with similar tasks.
A total of six annotators participated in the evaluation. In line with institutional requirements, each signed a Non-Disclosure Agreement (NDA), and a third-party company managed their compensation at standard hourly rates based on their location.

\noindent
\textbf{Annotation Agreement for Explanation.}
In Table \ref{tab:likert_score}, we summarize the annotation agreement scores of the explanations. We used 5-point Likert scale for various human evaluation metrics, including informativeness, clarity, plausibility, and faithfulness. We compute the average Likert scale value (from three annotators) for all evaluation metrics. We manually evaluated complete test sets for both ArMeme and Hateful meme except ArMeme with English explanation. For the later case we randomly selected 200 random examples. This decision was made to reduce the human annotation cost. The average agreement scores for the ArMeme dataset with Arabic explanations are $>4.5$ out of 5 indicating high agreement in all evaluation metrics. However, for the English explanations of ArMeme, the faithfulness and plausibility scores are relatively lower. To better understand this issue, we plan to conduct further annotation on the complete test set of explanations. For the Hateful Memes dataset, the average Likert scale agreement scores range from 4.04 to 4.10 out of 5.

In addition, we also computed the annotation agreement on ordinal scales by adopting the agreement index $r^*_{wg(j)}$ \cite{james1984estimating}, which compares observed variance in ratings to the maximum possible variance under complete disagreement. For each item, the agreement score is computed as:
$r^*_{wg(j)} = 1 - \frac{S_X^2}{\sigma^2_{\text{mv}}}$
where $S_X^2$ is the observed variance across annotators and $\sigma^2_{\text{mv}}$ is the maximum variance possible given the scale (computed as $\sigma^2_{\text{mv}} = 0.5(X_U^2 + X_L^2) - [0.5(X_U + X_L)]^2$, with $X_U = 5$ and $X_L = 1$ for a 5-point scale). The average agreement scores for ArMeme and Hateful memes are above 0.83 and 0.92, respectively, for all metrics. These values indicate strong agreement~\cite{o2017overview}. 

\begin{table}[h]
\centering
\setlength{\tabcolsep}{1pt} 
\scalebox{0.78}{%
\begin{tabular}{@{}lrrrr@{}}
\toprule
\multicolumn{1}{c}{Dataset} & \multicolumn{1}{c}{Faithfulness} & \multicolumn{1}{c}{Clarity} & \multicolumn{1}{c}{Plausiblity} & \multicolumn{1}{c}{Informative} \\ \midrule
ArMeme Ar expl. & 4.64 & 4.69 & 4.69 & 4.74 \\
ArMeme En expl. & 3.91 & 4.50 & 3.81 & 4.13 \\
Hateful meme & 4.01 & 4.18 & 4.04 & 4.10 \\ \bottomrule
\end{tabular}
}
\vspace{-0.3cm}
\caption{\textit{Average} Likert scale value for each human evaluation (annotation) metric across different sets of explanations.}
\label{tab:likert_score}
\vspace{-0.4cm}

\end{table}


\subsection{Basic Statistics}
Table \ref{tab:explation_stat} presents the basic statistics for both datasets. The average explanation length is 94 words for Arabic and 85 words for English. Notably, we instructed GPT-4o to generate explanations with fewer than 100 words. 
\section{Methodology}
\label{sec:experiments}


\subsection{Instructions Dataset}
Our approach follows the standard pipeline for aligning LLMs with user intentions and specific tasks through fine-tuning on representative data \cite{zhang2023instruction,kmainasi2024llamalens,hasan-etal-2025-nativqa}. 
This process typically involves curating and constructing instruction datasets that guide the model's behavior, ensuring it generates responses that align with the desired objectives. For our study, the responses include label and explanation. Hence, we created instruction format for both datasets.
For the ArMeme dataset, we replicated the experiments for both Arabic and English explanations.

\subsection{Model Selection}
As shown in Figure \ref{fig:vllm_exp_meme}, our first experimental phase involves model selection among several recent VLMs, including Llama-3.2 (11b) \cite{dubey2024llama}, Paligemma 2 (3b)~\cite{steiner2024paligemma}, Qwen2-vl~\cite{wang2024qwen2}, and Pixtral (12b)~\cite{agrawal2024pixtral}. The goal here is to select the VLM that maximizes \( p(l|i) \) for all image samples.

We evaluate the base models in a zero-shot setting and fine-tune them using an instruction-following paradigm. The instructions prompt the model to generate responses in the format \texttt{``Label: (class\_label)''}. We use a regex-based function to extract the predicted labels.



Note that this stage fine-tunes the models to predict class labels only, allowing us to verify whether they can handle multilingual inputs—especially in understanding Arabic text, cultural nuances, and image context. We do not ask the model to generate explanations here, as that is a more complex task and could affect their performance.


Based on the results reported in Tables \ref{tab:armeme} and \ref{tab:hateful_meme}, we selected Llama-3.2-vision-instruct (11b) for further training with explanations.



\subsection{\textit{Multi-Stage (MS)} Optimization Procedure} \label{sec:multi-stage}
To emphasize our contribution, we introduce a novel optimization procedure to train VLM with \memex{}, which decouples the classification and explanation generation tasks. 
While training both tasks in a single step may seem like an obvious choice, in practice, the two objectives produce \textit{conflicting gradient signals} due to the fundamental differences between the learning objectives. Classification requires precise mapping of multimodal cues to discrete labels, whereas explanation generation demands fluency and coherence in free-form natural language. Merging them too early can compromise the model's performance on both tasks. Conversely, training them completely separately risks overwriting knowledge of one task while learning the other. To address these challenges, we propose a two-stage optimization procedure that \textit{decouples} the learning objectives by first optimizing in the classification domain in \textit{stage-1}, followed by augmentation through explanation generation in \textit{stage-2}. The overall goal of this optimization procedure is to obtain an optimal VLM that maximizes \( p(l,e|i) \) for the training dataset (cf. \textit{Continual Training} in Figure \ref{fig:vllm_exp_meme}).

\paragraph{\textit{Problem Formulation.}}  
We formulate the joint optimization problem as follows: $L_{\mathrm{total}} = L_{\mathrm{classif}} + W_{\mathrm{expl}}\;\cdot\;L_{\mathrm{expl}},$
where \(L_{\mathrm{classif}}\) and \(L_{\mathrm{expl}}\) are the classification and explanation losses, respectively. \(W_{\mathrm{exp}}\) is a step‑function weight that switches from 0 to 1 between the stages. This formulation draws on the principles from \emph{Domain Adaptation} and \emph{Task‑Incremental Learning} (\citealp{van2022three}) with the aim to: (a) \textit{Isolate classification learning} and avoid conflicting updates and  (b) \textit{Prevent catastrophic forgetting} by gradually integrating the explanation objective.
\noindent
\paragraph{\textit{Stage 1 - Classification Fine-Tuning.}} 
We set \(W_{\mathrm{exp}}=0\) and optimize exclusively on \(L_{\mathrm{classif}}\). It adapts the pretrained VLM to the domain of hateful/propagandistic content, establishing a strong feature backbone for accurate label prediction.
\noindent
\paragraph{\textit{Stage 2 - Joint Classification \& Explanation Fine‑Tuning.}}  
We set \(W_{\mathrm{exp}}=1\) to optimize $L_{\mathrm{classif}} + L_{\mathrm{explanation}}$.
After obtaining the domain-adapted backbone from Stage 1, the model proceeds to learn how to generate coherent, contextually grounded explanations alongside accurate classifications. This stepwise integration ensures that the model preserves its classification capabilities and avoids catastrophic forgetting, while developing proficiency in natural language reasoning.

By decoupling and then recombining these objectives in a straightforward two-phase procedure, our multi-stage (MS) optimization presents an easily implementable extension to standard VLM training pipelines.
%
To validate its effectiveness, we compare it against a single-stage (SS) fine-tuning baseline, where the model is directly trained on the label-with-explanation dataset. Our ablation studies (detailed in Section \ref{label:results}) demonstrate that the proposed multi-stage approach significantly outperforms the single-stage strategy.

\begin{table}[h]
\centering
\setlength{\tabcolsep}{3pt} 
\scalebox{0.85}{%
\begin{tabular}{@{}llrrr@{}}
\toprule
\multicolumn{1}{c}{\textbf{Model}} & \multicolumn{1}{c}{\textbf{Setup}} & \multicolumn{1}{c}{\textbf{Acc (\%)}} & \multicolumn{1}{c}{\textbf{W-F1}} & \multicolumn{1}{c}{\textbf{M-F1}} \\ \midrule
\cite{alam-etal-2024-armeme}  & Qarib & 69.7 & 0.690 & 0.551 \\
\cite{alam-etal-2024-armeme} & mBERT & 70.7 & 0.675 & 0.487 \\
Llama-3.2 (11b) & Base & 13.4 & 0.172 & 0.113 \\
Llama-3.2 (11b) & FT & 68.0 & 0.665 & \textbf{0.452} \\
Paligemma2 (3b) & Base & 15.3 & 0.090 & 0.080 \\
Paligemma2 (3b) & FT & 65.9 & 0.524 & 0.200 \\
Qwen2V (7b) & Base & 63.1 & 0.550 & 0.242 \\
Qwen2V (7b) & FT & 72.2 & 0.686 & 0.440 \\
Pixtral (12b) & Base & 14.6 & 0.177 & 0.133 \\
Pixtral (12b) & FT & 70.8 & 0.636 & 0.377 \\ 
\bottomrule
\end{tabular}
}
\vspace{-0.2cm}
\caption{Results for ArMeme. FT: Fine-tuned. Qarib~\cite{abdelali2021pretraining} is a Arabic BERT (text only). mBERT - multilingual BERT (text only).}
\label{tab:armeme}
\end{table}

\begin{table}[h]
\centering
\setlength{\tabcolsep}{3pt} 
\scalebox{0.85}{%
\begin{tabular}{@{}llrrr@{}}
\toprule
\multicolumn{1}{c}{\textbf{Model}} & \multicolumn{1}{c}{\textbf{Setup}} & \multicolumn{1}{c}{\textbf{Acc (\%)}} & \multicolumn{1}{c}{\textbf{W-F1}} & \multicolumn{1}{c}{\textbf{M-F1}} \\ \midrule
\cite{kiela2020hateful} &  & 69.47±2.06 &  &  \\  
\cite{cao-etal-2022-prompting} &  & 72.98±1.09 &  &  \\ 
\cite{burbi2023mapping} &  & 77.70 &  &  \\ 
Llama-3.2 (11b) & Base & 66.1 & 0.650 & 0.618 \\
Llama-3.2 (11b) & FT & 77.7 & 0.770 & \textbf{0.748} \\
Paligemma2 (3b) & Base & 35.2 & 0.277 & 0.217 \\
Paligemma2 (3b) & FT & 69.2 & 0.664 & 0.623 \\
Qwen2V (7b) & Base & 66.4 & 0.669 & 0.442 \\
Qwen2V (7b) & FT & 77.9 & 0.773 & 0.753 \\
Pixtral (12b) & Base & 66.7 & 0.667 & 0.430 \\
Pixtral (12b) & FT & 77.2 & 0.766 & \underline{0.746} \\ \bottomrule
\end{tabular}
}
\vspace{-0.2cm}
\caption{Results for Hateful meme. FT: Fine-tuned}
\label{tab:hateful_meme}
\vspace{-0.2cm}
\end{table}

\subsection{Training Setup}

Our fine-tuning experiments utilize QLoRA~\cite{dettmers2023qloraefficientfinetuningquantized}, which combines INT4 quantization with parameter-efficient fine-tuning through Low-Rank Adaptation (LoRA)~\cite{hu2021lora}. In our setup, the base model is quantized to 4-bit precision, with LoRA updates applied to a subset of the model parameters. 
This approach was selected to address computational and memory resource constraints. 
We adapted all relevant submodules (vision, language, attention, and MLP layers) with a LoRA rank of 16, an alpha of 16, and no dropout. 
We used a per-device batch size of 2 with gradient accumulation over 4 steps, optimizing with AdamW, a weight decay of 0.01, and a linear scheduler with 5 warmup steps.

\paragraph{Single-Stage (SS) Baseline.}
We implement a baseline where models are fine-tuned end-to-end using a unified classification and classification-with-explanation objectives. Model selection is performed on the development set, and final results are reported on the test set using the best-performing checkpoint.

\paragraph{Multi-Stage (MS) Curriculum.}
We design a staged curriculum to progressively instill multimodal reasoning capabilities into the model. In the first stage, the model is trained exclusively on the classification dataset, and intermediate checkpoints are evaluated on the development set. The best checkpoint from this stage is then used to initialize the second stage, where the model is further optimized with the classification-with-explanation objective. Hyperparameter exploration revealed that lowering the learning rate while increasing the number of training epochs in the second stage yields performance gains in our setup.

\noindent
\subsection{Evaluation Setup and Metrics}
The results for pretrained models, fine-tuned models, and baselines are summarized in Tables \ref{tab:armeme} and \ref{tab:hateful_meme}.
%
For performance measurement across different experimental settings, we compute accuracy, weighted F$_1$, and macro-F$_1$. 
We evaluate the model's explanation performance on the test set using semantic similarity-based metric, measured by the F$_1$ score within BERTScore~\citep{zhangbertscore}. This score is computed using contextual embeddings extracted from pre-trained BERT models. To enhance accuracy, we utilize language-specific transformer models for embedding extraction. For Arabic we use AraBERT (v2)~\cite{baly2020arabert} model and for English we use bert-base-uncased model~\cite{devlin2019bert}. In addition, we also compute BLEU and METEOR scores.

\section{Experimental Results and Discussion}
\label{label:results}


This section first presents competitive results among our proposed method and the state-of-the-art approaches. Next, it briefly analyzes and investigates the proposed method to validate and highlight the core contributions of this research. 
%
To increase clarity, we emphasize that all the results discussed in this section are based on an experimental setup that evaluates models using the \memex{} dataset, which includes both classification and explanation for training. This differs from the results presented in the previous section, in Tables \ref{tab:armeme} and \ref{tab:hateful_meme}, where we evaluated and compared models trained on the original datasets, which include only classification labels. Therefore, none of the results discussed in this section should be compared to those presented in Tables \ref{tab:armeme} and \ref{tab:hateful_meme}.

\paragraph{Comparison with State‑of‑the‑Art.}
Table~\ref{tab:sota_comparison} presents the comparison.
On the ArMeme dataset, our proposed approach (\textit{Llama MS}) achieves the highest accuracy at 72.1\% and the best weighted F1 score at 0.699, with Qarib and mBERT trailing behind. Although Qarib achieves the highest macro F1 score (0.551), \textit{Llama MS} remains competitive with a macro F1 of 0.536. Importantly, our approach stands out by providing explanations that add significant value.
On the Hateful Meme dataset, \textit{Llama MS} surpasses the state-of-the-art by achieving the best performance: an accuracy of 79.9\%, a weighted F1 score of 0.802, and a macro F1 score of 0.792.
These results clearly highlight the advantages of our explainability-enhanced dataset and the effectiveness of our proposed optimization procedure for both classification and explanation-generation tasks.

\begin{table}[t]
\centering
\setlength{\tabcolsep}{3pt} 
\scalebox{0.80}{%
\begin{tabular}{@{}llrrr@{}}
\toprule
\multicolumn{1}{c}{\textbf{Model}} & \multicolumn{1}{c}{\textbf{Setup}} & \multicolumn{1}{c}{\textbf{Acc(\%)}} & \multicolumn{1}{c}{\textbf{W-F1}} & \multicolumn{1}{c}{\textbf{M-F1}} \\ \midrule
\multicolumn{5}{c}{\textbf{ArMeme}} \\ \midrule
\cite{alam-etal-2024-armeme} & Qarib & 69.7 & 0.690 & 0.551 \\
\cite{alam-etal-2024-armeme} & mBERT & 70.7 & 0.675 & 0.487 \\
\cite{alam-etal-2024-armeme} & ResNet50  & 66.0 & 0.637 & 0.434 \\
Llama MS & FT & 72.1 & 0.699 & 0.536 \\
Llama (Ar-Exp) MS & FT & 72.0 & 0.696 & 0.499 \\ \midrule
\multicolumn{5}{c}{\textbf{Hateful Meme}} \\ \midrule
\cite{kiela2020hateful} &  & 69.47±2.06 &  &  \\
\cite{cao-etal-2022-prompting} &  & 72.98±1.09 &  &  \\
\cite{burbi2023mapping} &  & 77.70 &  &  \\ 
Llama MS & FT & 79.9 & 0.802 & 0.792 \\ \bottomrule
\end{tabular}
}
\vspace{-0.2cm}
\caption{Comparison with SOTA and our results (\textit{Llama MS and Llama (Ar-Exp) MS}). ResNet50~\cite{he2016deep} is an image only model. MS: Multi-stage \textit{(Our approach)}.}
\label{tab:sota_comparison}
\vspace{-0.3cm}
\end{table}


\begin{table}[htb!]
\centering
\setlength{\tabcolsep}{2.5pt} 
\scalebox{0.72}{%
\begin{tabular}{@{}llrrrrrr@{}}
\toprule
\multicolumn{1}{c}{\textbf{Model}} & \multicolumn{1}{c}{\textbf{Setup}} & \multicolumn{1}{c}{\textbf{Acc (\%)}} & \multicolumn{1}{c}{\textbf{W-F1}} & \multicolumn{1}{c}{\textbf{M-F1}} & \multicolumn{1}{c}{\textbf{BS}} & \multicolumn{1}{c}{\textbf{BL}}& \multicolumn{1}{c}{\textbf{M}}\\  \midrule
\multicolumn{8}{c}{\textbf{ArMeme}} \\  \midrule
Llama & Base & 12.7 & 0.165 & 0.105 & 0.61 & 0.24 & 0.17\\
Llama SS & FT & 68.2 & 0.584 & 0.257 & 0.70 & 0.56 & 0.36 \\
Llama MS & FT & 72.1 & 0.699 & 0.536 & 0.70 & 0.57 & 0.35 \\
Llama Ar-Exp & Base & 19.0 & 0.246 & 0.125 & 0.58 & 0.12 & 0.09 \\ 
Llama MS Ar-Exp & FT & 72.0 & 0.696 & 0.499 & 0.72 & 0.55 & 0.29 \\ \midrule
\multicolumn{8}{c}{\textbf{Hateful Meme}} \\  \midrule
Llama & Base & 65.2 & 0.615 & 0.567 & 0.661 & 0.35 & 0.23 \\
Llama SS & FT & 75.9 & 0.760 & 0.745 & 0.767 & 0.65 & 0.47 \\
Llama MS & FT & 79.9 & 0.802 & 0.792 & 0.777 & 0.67 & 0.49 \\ \bottomrule
\end{tabular}
}
\vspace{-0.2cm}
\caption{Results with ArMeme and Hateful meme classification and explanation generation. Llama: Llama-3.2 (11b), BS: BERTScore, BL: BLEU, M: METEOR. SS: Single-stage, MS: Multi-stage. Ar-Exp: Model trained with Arabic explanation. W-F1: Weighted F1, M-F1: macro-F1}
\label{tab:results_expl}
\vspace{-0.3cm}
\end{table}


\paragraph{Ablation Study on Different Model Training Settings.}
Table~\ref{tab:results_expl} presents classification and explanation-generation results on the \emph{ArMeme} and \emph{Hateful Meme} datasets across different settings: (a) \textit{Llama} represents the performance of the base model without fine-tuning; (b) \textit{Llama SS} represents the performance after single-stage fine-tuning; and (c) \textit{Llama MS} represents the performance after completing the second stage of our proposed multi-stage optimization. The extension \textit{Ar-Exp} after a setting name indicates that the model was trained using Arabic explanations. Based on these settings, we analyze the results from various perspectives:
\noindent
\textit{(a)} \textbf{Base vs.~FT}: demonstrates the performance difference between the same model with and without fine-tuning (FT); \\
\textit{(b)} \textbf{Single-stage (SS) vs.~Multi-stage (MS)}: highlights the necessity and benefits of the proposed optimization procedure; and \\
\textit{(c)} \textbf{Eng-Exp vs.~Ar-Exp}: showcases the multilingual capability of the selected VLM. 
%

First, we compare the \textbf{Base vs.~FT} setup in Table~\ref{tab:results_expl}, from which it is evident that the FT model significantly outperforms the baseline. For example, on the ArMeme dataset, while the baseline (\textit{Llama}) achieves an accuracy of 12.7\%, the proposed fine-tuning (\textit{Llama-MS}) boosts it to 72.1\%. Similarly, on the Hateful Meme dataset, fine-tuning improves the base accuracy from 65.2\% (\textit{Llama}) to 79.9\% (\textit{Llama-MS}). We observe similar improvements in the F1 metrics for classification and BERTScore for explanation quality. These significant performance gains \textit{validate our approach of fine-tuning the base models with the explainability enhanced dataset}, demonstrating its efficacy for the meme classification and explanation generation tasks.
%

Next, we compare the \textbf{SS vs.~MS} setup in Table~\ref{tab:results_expl}, which reveals that multi-stage (MS) fine-tuning further enhances performance over the classification single-stage approach. For example, on the ArMeme dataset, the accuracy increased from 68.2\% (\textit{Llama-SS}) to 72.1\%  (\textit{Llama-MS}), the weighted F1 increased from 0.584 (\textit{Llama-SS}) to 0.699  (\textit{Llama-MS}), the macro F1 increased significantly from 0.257 (\textit{Llama-SS}) to 0.536 (\textit{Llama-MS}), and the BERTScore for Arabic explanation increased significantly from 0.58 (\textit{Llama-SS}) to 0.72 (\textit{Llama-MS}). A similar trend is observed on the Hateful Meme dataset, where additional fine-tuning iterations yield more robust classification (approximately 4\% improvement) and enhanced explanation quality. These performance gains \textit{validate our proposed multi-stage optimization procedure} to further refine the VLMs.
%

\paragraph{Assessment of Multilingual Capability.}
We compare \textbf{\textit{Llama MS - FT}} with \textbf{\textit{Llama MS Ar-Exp - FT}} in Table~\ref{tab:results_expl}, which shows that fine-tuning using explanations generated in both languages yields comparable outcomes. This \textit{validates the multilingual capability of our empirically chosen VLM} for the target task and helps users understand multilingual content without fluency in that language. For example, our model allows an English speaker to analyze Arabic memes and receive explanations in English.

\paragraph{Annotation Agreement: Additional Analysis}
\label{sec:app_annotation_agreement}

We report human evaluations of the generated explanations from two models: (i) Llama MS (ArMeme) and (ii) Llama MS (Hateful Memes), as detailed in Table \ref{tab:likert_score_memeintel_models}. Due to resource constraints, the manual evaluation was conducted on a representative sample of 100 instances. For the ArMeme dataset, the scores across all evaluation metrics range from 4.15 to 4.74 out of 5, while for the Hateful Memes dataset, the scores range from 4.41 to 4.54, indicating consistently strong performance across both datasets.


\begin{table}[h]
\centering
\setlength{\tabcolsep}{1pt} 
\scalebox{0.78}{%
\begin{tabular}{@{}lrrrr@{}}
\toprule
\multicolumn{1}{c}{\textbf{Dataset}} & \multicolumn{1}{c}{\textbf{Faithfulness}} & \multicolumn{1}{c}{\textbf{Clarity}} & \multicolumn{1}{c}{\textbf{Plausiblity}} & \multicolumn{1}{c}{\textbf{Informative}} \\ \midrule
ArMeme & 4.63 & 4.74 & 4.56 & 4.15 \\
Hateful meme & 4.41 & 4.44 & 4.43 & 4.54 \\ \bottomrule
\end{tabular}
}
\vspace{-0.3cm}
\caption{\textit{Average} Likert scale value for each human evaluation (annotation) metric for the explanations of different datasets.}
\label{tab:likert_score_memeintel_models}
\end{table}

\noindent
\paragraph{Error Analysis.}
To better understand the model's reasoning capabilities and failure modes, we conduct an error analysis contrasting correct and incorrect label predictions, as well as comparing single-stage and multistage training paradigms. Our analysis reveals that while the model often aligns well with human explanations in correctly classified instances, it struggles with nuanced cases involving humor or implicit context. Furthermore, multistage training consistently outperforms single-stage training in label accuracy by better integrating textual and visual cues while providing explanation. See Section~\ref{ssec:app_error_analysis} for detailed examples and discussion.

\section{Conclusions and Future Work}
\label{sec:conclusions}
In this study, we introduce a \memex{} dataset for propagandistic and hateful meme detection and natural explanation generation, making it the \textit{first} resource of its kind. To address both detection and explanation generation tasks and ensure efficient VLMs model training on this dataset, we also propose a multi-stage optimization procedure.  
To evaluate the multilingual capability of the model, we developed Arabic and English explanations for Arabic memes. The inclusion of English explanations benefits non-Arabic speakers, whereas providing explanations in the native language ensures that cultural nuances are accurately conveyed. With our training procedure, we demonstrate improved detection performance for both \textit{ArMeme} and \textit{hateful} memes. The higher performance of explanation generation further demonstrates the efficacy of our multi-stage training approach.  

Moreover, the proposed multi-stage scheme is agnostic to specific VLM architectures and applicable in contexts where tasks impose divergent gradient demands, thereby offering a versatile framework for multi-task vision–language learning.
%

We foresee several future directions to extend this research and explore the following: \textit{(a)} training the model with additional data through data augmentation, which could help it become an instruction-generalized model and potentially enhance its performance further; \textit{(b)} incorporating pseudo and self-labeled data using an active learning procedure to incrementally improve the model's capabilities; and \textit{(c)} developing a task-generalized model that addresses multiple tasks.

\section{Limitations}
Due to the complex nature of manual explanation creation, we have relied on GPT-4o for explanation generation. To ensure the reliability of the explanation we have manually evaluated in four criteria such as informativeness, clarity, plausiblity, and faithfulness on a small sample for each set of explanation. The preliminary evaluation scores suggest that we can rely on the gold explanation as the reference. As a part of ongoing work we plan to conduct manual evaluation on a larger set. An important aspect of the ArMeme dataset is that it is highly imbalanced, which affects overall performance. One possible approach to address this issue is to increase the number of memes labeled as propaganda, other, and not-meme. This can be achieved through data augmentation or by collecting additional memes.

\section*{Ethics and Broader Impact}
We extended existing datasets by adding explanations. To the best of our knowledge, the dataset does not contain any personally identifiable information, making privacy risks nonexistent. Regarding the explanations, we provided clear annotation instructions and cautioned annotators that some memes might be offensive. It is important to note that annotations are inherently subjective, which can introduce biases into the overall evaluation results. We encourage researchers and users of this dataset to remain critical when developing models or conducting further research. Models built using this dataset could be highly valuable for fact-checkers, journalists, and social media platforms.

\section*{Acknowledgments}
The work of F. Alam is supported by the NPRP grant 14C-0916-210015 from the Qatar National Research Fund part of Qatar Research Development and Innovation Council (QRDI). The findings achieved herein are solely the responsibility of the authors.

\bibliography{bibliography/main}

\begin{thebibliography}{56}
\providecommand{\natexlab}[1]{#1}

\bibitem[{Abdelali et~al.(2021)Abdelali, Hassan, Mubarak, Darwish, and Samih}]{abdelali2021pretraining}
Ahmed Abdelali, Sabit Hassan, Hamdy Mubarak, Kareem Darwish, and Younes Samih. 2021.
\newblock \href {https://arxiv.org/abs/2102.10684} {Pre-training bert on arabic tweets: Practical considerations}.

\bibitem[{Agarwal et~al.(2024)Agarwal, Tanneru, and Lakkaraju}]{agarwal_faithfulness_2024}
Chirag Agarwal, Sree~Harsha Tanneru, and Himabindu Lakkaraju. 2024.
\newblock \href {https://doi.org/10.48550/arXiv.2402.04614} {Faithfulness vs. plausibility: On the (un)reliability of explanations from large language models}.
\newblock \emph{arXiv preprint arXiv:2402.04614}, 2402.04614.

\bibitem[{Agrawal et~al.(2024)Agrawal, Antoniak, Hanna, Bout, Chaplot, Chudnovsky, Costa, De~Monicault, Garg, Gervet et~al.}]{agrawal2024pixtral}
Pravesh Agrawal, Szymon Antoniak, Emma~Bou Hanna, Baptiste Bout, Devendra Chaplot, Jessica Chudnovsky, Diogo Costa, Baudouin De~Monicault, Saurabh Garg, Theophile Gervet, et~al. 2024.
\newblock Pixtral 12b.
\newblock \emph{arXiv preprint arXiv:2410.07073}.

\bibitem[{Alam et~al.(2024{\natexlab{a}})Alam, Biswas, Shah, Zaghouani, and Mikros}]{alam2024propaganda}
Firoj Alam, Md~Rafiul Biswas, Uzair Shah, Wajdi Zaghouani, and Georgios Mikros. 2024{\natexlab{a}}.
\newblock Propaganda to hate: A multimodal analysis of arabic memes with multi-agent llms.
\newblock In \emph{International Conference on Web Information Systems Engineering}, pages 380--390. Springer.

\bibitem[{Alam et~al.(2022)Alam, Cresci, Chakraborty, Silvestri, Dimitrov, Da~San~Martino, Shaar, Firooz, and Nakov}]{alam2022survey}
Firoj Alam, Stefano Cresci, Tanmoy Chakraborty, Fabrizio Silvestri, Dimiter Dimitrov, Giovanni Da~San~Martino, Shaden Shaar, Hamed Firooz, and Preslav Nakov. 2022.
\newblock A survey on multimodal disinformation detection.
\newblock In \emph{Proceedings of the 29th International Conference on Computational Linguistics}, pages 6625--6643.

\bibitem[{Alam et~al.(2024{\natexlab{b}})Alam, Hasnat, Ahmad, Hasan, and Hasanain}]{alam-etal-2024-armeme}
Firoj Alam, Abul Hasnat, Fatema Ahmad, Md.~Arid Hasan, and Maram Hasanain. 2024{\natexlab{b}}.
\newblock \href {https://doi.org/10.18653/v1/2024.emnlp-main.1173} {{A}r{M}eme: Propagandistic content in {A}rabic memes}.
\newblock In \emph{Proceedings of the 2024 Conference on Empirical Methods in Natural Language Processing}, pages 21071--21090, Miami, Florida, USA. Association for Computational Linguistics.

\bibitem[{Antoun et~al.(2020)Antoun, Baly, and Hajj}]{baly2020arabert}
Wissam Antoun, Fady Baly, and Hazem Hajj. 2020.
\newblock {AraBERT}: Transformer-based model for {Arabic} language understanding.
\newblock In \emph{Proceedings of the 4th Workshop on Open-Source Arabic Corpora and Processing Tools, with a Shared Task on Offensive Language Detection}, pages 9--15.

\bibitem[{Barr{\'o}n-Cedeno et~al.(2019)Barr{\'o}n-Cedeno, Jaradat, Da~San~Martino, and Nakov}]{BARRONCEDENO20191849}
Alberto Barr{\'o}n-Cedeno, Israa Jaradat, Giovanni Da~San~Martino, and Preslav Nakov. 2019.
\newblock Proppy: Organizing the news based on their propagandistic content.
\newblock \emph{Information Processing \& Management}, 56(5):1849--1864.

\bibitem[{Burbi et~al.(2023)Burbi, Baldrati, Agnolucci, Bertini, and Del~Bimbo}]{burbi2023mapping}
Giovanni Burbi, Alberto Baldrati, Lorenzo Agnolucci, Marco Bertini, and Alberto Del~Bimbo. 2023.
\newblock Mapping memes to words for multimodal hateful meme classification.
\newblock In \emph{Proceedings of the IEEE/CVF International Conference on Computer Vision}, pages 2832--2836.

\bibitem[{Cao et~al.(2023)Cao, Hee, Kuek, Chong, Lee, and Jiang}]{Rui2023}
Rui Cao, Ming~Shan Hee, Adriel Kuek, Wen-Haw Chong, Roy Ka-Wei Lee, and Jing Jiang. 2023.
\newblock \href {https://doi.org/10.1145/3581783.3612498} {Pro-cap: Leveraging a frozen vision-language model for hateful meme detection}.
\newblock In \emph{Proceedings of the 31st ACM International Conference on Multimedia}, MM '23, page 5244–5252, New York, NY, USA. Association for Computing Machinery.

\bibitem[{Cao et~al.(2022)Cao, Lee, Chong, and Jiang}]{cao-etal-2022-prompting}
Rui Cao, Roy Ka-Wei Lee, Wen-Haw Chong, and Jing Jiang. 2022.
\newblock \href {https://doi.org/10.18653/v1/2022.emnlp-main.22} {Prompting for multimodal hateful meme classification}.
\newblock In \emph{Proceedings of the 2022 Conference on Empirical Methods in Natural Language Processing}, pages 321--332, Abu Dhabi, United Arab Emirates. Association for Computational Linguistics.

\bibitem[{Chen et~al.(2024)Chen, Zhao, Piao, Ding, and Cui}]{chen2024multimodal}
Pengyuan Chen, Lei Zhao, Yangheran Piao, Hongwei Ding, and Xiaohui Cui. 2024.
\newblock Multimodal visual-textual object graph attention network for propaganda detection in memes.
\newblock \emph{Multimedia Tools and Applications}, 83(12):36629--36644.

\bibitem[{Da~San~Martino et~al.(2019)Da~San~Martino, Yu, Barr{\'o}n-Cede{\~n}o, Petrov, and Nakov}]{da-san-martino-etal-2019-fine}
Giovanni Da~San~Martino, Seunghak Yu, Alberto Barr{\'o}n-Cede{\~n}o, Rostislav Petrov, and Preslav Nakov. 2019.
\newblock Fine-grained analysis of propaganda in news article.
\newblock In \emph{Proceedings of the 2019 Conference on Empirical Methods in Natural Language Processing and the 9th International Joint Conference on Natural Language Processing (EMNLP-IJCNLP)}, Hong Kong, China.

\bibitem[{Dettmers et~al.(2023)Dettmers, Pagnoni, Holtzman, and Zettlemoyer}]{dettmers2023qloraefficientfinetuningquantized}
Tim Dettmers, Artidoro Pagnoni, Ari Holtzman, and Luke Zettlemoyer. 2023.
\newblock \href {https://arxiv.org/abs/2305.14314} {Qlora: Efficient finetuning of quantized llms}.
\newblock \emph{Preprint}, arXiv:2305.14314.

\bibitem[{Devlin et~al.(2019)Devlin, Chang, Lee, and Toutanova}]{devlin2019bert}
Jacob Devlin, Ming-Wei Chang, Kenton Lee, and Kristina Toutanova. 2019.
\newblock {BERT}: Pre-training of deep bidirectional transformers for language understanding.
\newblock In \emph{Proceedings of the 2019 Conference of the North American Chapter of the Association for Computational Linguistics: Human Language Technologies}, NAACL-HLT~'19, Minneapolis, Minnesota, USA.

\bibitem[{Dimitrov et~al.(2024)Dimitrov, Alam, Hasanain, Hasnat, Silvestri, Nakov, and Martino}]{dimitrov2024semeval}
Dimitar Dimitrov, Firoj Alam, Maram Hasanain, Abul Hasnat, Fabrizio Silvestri, Preslav Nakov, and Giovanni Da~San Martino. 2024.
\newblock Semeval-2024 task 4: Multilingual detection of persuasion techniques in memes.
\newblock In \emph{Proceedings of the 2024 Annual Conference of the North American Chapter of the Association for Computational Linguistics}.

\bibitem[{Dimitrov et~al.(2021{\natexlab{a}})Dimitrov, Bin~Ali, Shaar, Alam, Silvestri, Firooz, Nakov, and Da~San~Martino}]{ACL2021:propaganda:memes}
Dimitar Dimitrov, Bishr Bin~Ali, Shaden Shaar, Firoj Alam, Fabrizio Silvestri, Hamed Firooz, Preslav Nakov, and Giovanni Da~San~Martino. 2021{\natexlab{a}}.
\newblock Detecting propaganda techniques in memes.
\newblock In \emph{ACL-IJCNLP}.

\bibitem[{Dimitrov et~al.(2021{\natexlab{b}})Dimitrov, Bin~Ali, Shaar, Alam, Silvestri, Firooz, Nakov, and Da~San~Martino}]{dimitrov-etal-2021-semeval}
Dimitar Dimitrov, Bishr Bin~Ali, Shaden Shaar, Firoj Alam, Fabrizio Silvestri, Hamed Firooz, Preslav Nakov, and Giovanni Da~San~Martino. 2021{\natexlab{b}}.
\newblock \href {https://doi.org/10.18653/v1/2021.semeval-1.7} {{S}em{E}val-2021 task 6: Detection of persuasion techniques in texts and images}.
\newblock In \emph{Proceedings of the 15th International Workshop on Semantic Evaluation (SemEval-2021)}, pages 70--98, Online. Association for Computational Linguistics.

\bibitem[{Dubey et~al.(2024)Dubey, Jauhri, Pandey, Kadian, Al-Dahle, Letman, Mathur, Schelten, Yang, Fan et~al.}]{dubey2024llama}
Abhimanyu Dubey, Abhinav Jauhri, Abhinav Pandey, Abhishek Kadian, Ahmad Al-Dahle, Aiesha Letman, Akhil Mathur, Alan Schelten, Amy Yang, Angela Fan, et~al. 2024.
\newblock The llama 3 herd of models.
\newblock \emph{arXiv preprint arXiv:2407.21783}.

\bibitem[{Habernal et~al.(2017)Habernal, Hannemann, Pollak, Klamm, Pauli, and Gurevych}]{Habernal.et.al.2017.EMNLP}
Ivan Habernal, Raffael Hannemann, Christian Pollak, Christopher Klamm, Patrick Pauli, and Iryna Gurevych. 2017.
\newblock Argotario: Computational argumentation meets serious games.
\newblock In \emph{Proceedings of the 2017 Conference on Empirical Methods in Natural Language Processing: System Demonstrations}, EMNLP~'17, pages 7--12, Copenhagen, Denmark.

\bibitem[{Habernal et~al.(2018)Habernal, Pauli, and Gurevych}]{Habernal2018b}
Ivan Habernal, Patrick Pauli, and Iryna Gurevych. 2018.
\newblock Adapting serious game for fallacious argumentation to {G}erman: Pitfalls, insights, and best practices.
\newblock In \emph{LREC}. European Language Resources Association (ELRA).

\bibitem[{Hasan et~al.(2025)Hasan, Hasanain, Ahmad, Laskar, Upadhyay, Sukhadia, Kutlu, Chowdhury, and Alam}]{hasan-etal-2025-nativqa}
Md.~Arid Hasan, Maram Hasanain, Fatema Ahmad, Sahinur~Rahman Laskar, Sunaya Upadhyay, Vrunda~N Sukhadia, Mucahid Kutlu, Shammur~Absar Chowdhury, and Firoj Alam. 2025.
\newblock \href {https://doi.org/10.18653/v1/2025.findings-acl.770} {{N}ativ{QA}: Multilingual culturally-aligned natural query for {LLM}s}.
\newblock In \emph{Findings of the Association for Computational Linguistics: ACL 2025}, pages 14886--14909, Vienna, Austria. Association for Computational Linguistics.

\bibitem[{Hasanain et~al.(2023)Hasanain, Alam, Mubarak, Abdaljalil, Zaghouani, Nakov, Da~San~Martino, and Freihat}]{hasanain-etal-2023-araieval}
Maram Hasanain, Firoj Alam, Hamdy Mubarak, Samir Abdaljalil, Wajdi Zaghouani, Preslav Nakov, Giovanni Da~San~Martino, and Abed Freihat. 2023.
\newblock \href {https://doi.org/10.18653/v1/2023.arabicnlp-1.44} {{A}r{AIE}val shared task: Persuasion techniques and disinformation detection in {A}rabic text}.
\newblock In \emph{Proceedings of ArabicNLP 2023}, pages 483--493, Singapore (Hybrid). Association for Computational Linguistics.

\bibitem[{Hasanain et~al.(2024)Hasanain, Hasan, Ahmed, Suwaileh, Biswas, Zaghouani, and Alam}]{araieval:arabicnlp2024-overview}
Maram Hasanain, Md.~Arid Hasan, Fatema Ahmed, Reem Suwaileh, Md.~Rafiul Biswas, Wajdi Zaghouani, and Firoj Alam. 2024.
\newblock Araieval shared task: Propagandistic techniques detection in unimodal and multimodal arabic content.
\newblock In \emph{Proceedings of the Second Arabic Natural Language Processing Conference (ArabicNLP 2024)}, Bangkok. Association for Computational Linguistics.

\bibitem[{He et~al.(2016)He, Zhang, Ren, and Sun}]{he2016deep}
Kaiming He, Xiangyu Zhang, Shaoqing Ren, and Jian Sun. 2016.
\newblock Deep residual learning for image recognition.
\newblock In \emph{Proceedings of the IEEE conference on computer vision and pattern recognition}, CVPR~'16, pages 770--778. IEEE.

\bibitem[{Hee et~al.(2023)Hee, Chong, and Lee}]{hee2023decoding}
Ming~Shan Hee, Wen-Haw Chong, and Roy Ka-Wei Lee. 2023.
\newblock Decoding the underlying meaning of multimodal hateful memes.
\newblock In \emph{Proceedings of the Thirty-Second International Joint Conference on Artificial Intelligence}, pages 5995--6003.

\bibitem[{Hee et~al.(2022)Hee, Lee, and Chong}]{hee2022explaining}
Ming~Shan Hee, Roy Ka-Wei Lee, and Wen-Haw Chong. 2022.
\newblock On explaining multimodal hateful meme detection models.
\newblock In \emph{Proceedings of the ACM Web Conference 2022}, pages 3651--3655.

\bibitem[{Herm(2023)}]{herm2023impact}
Lukas-Valentin Herm. 2023.
\newblock Impact of explainable ai on cognitive load: Insights from an empirical study.
\newblock \emph{arXiv preprint arXiv:2304.08861}.

\bibitem[{Hossain et~al.(2022)Hossain, Sharif, and Hoque}]{hossain-etal-2022-mute}
Eftekhar Hossain, Omar Sharif, and Mohammed~Moshiul Hoque. 2022.
\newblock \href {https://doi.org/10.18653/v1/2022.aacl-srw.5} {{MUTE}: A multimodal dataset for detecting hateful memes}.
\newblock In \emph{Proceedings of the 2nd Conference of the Asia-Pacific Chapter of the Association for Computational Linguistics and the 12th International Joint Conference on Natural Language Processing: Student Research Workshop}, pages 32--39, Online. Association for Computational Linguistics.

\bibitem[{Hu et~al.(2022)Hu, Shen, Wallis, Allen{-}Zhu, Li, Wang, Wang, and Chen}]{hu2021lora}
Edward~J. Hu, Yelong Shen, Phillip Wallis, Zeyuan Allen{-}Zhu, Yuanzhi Li, Shean Wang, Lu~Wang, and Weizhu Chen. 2022.
\newblock Lora: Low-rank adaptation of large language models.
\newblock In \emph{The Tenth International Conference on Learning Representations, {ICLR} 2022, Virtual Event, April 25-29, 2022}. OpenReview.net.

\bibitem[{Huang et~al.(2023)Huang, Kwak, and An}]{huang_chain_2023}
Fan Huang, Haewoon Kwak, and Jisun An. 2023.
\newblock Chain of explanation: New prompting method to generate quality natural language explanation for implicit hate speech.
\newblock In \emph{Companion Proceedings of the ACM Web Conference 2023}, pages 90--93.

\bibitem[{James et~al.(1984)James, Demaree, and Wolf}]{james1984estimating}
Lawrence~R James, Robert~G Demaree, and Gerrit Wolf. 1984.
\newblock Estimating within-group interrater reliability with and without response bias.
\newblock \emph{Journal of applied psychology}, 69(1):85.

\bibitem[{Kiela et~al.(2020)Kiela, Firooz, Mohan, Goswami, Singh, Ringshia, and Testuggine}]{kiela2020hateful}
Douwe Kiela, Hamed Firooz, Aravind Mohan, Vedanuj Goswami, Amanpreet Singh, Pratik Ringshia, and Davide Testuggine. 2020.
\newblock The hateful memes challenge: Detecting hate speech in multimodal memes.
\newblock \emph{Advances in neural information processing systems}, 33:2611--2624.

\bibitem[{Kmainasi et~al.(2024)Kmainasi, Shahroor, Hasanain, Laskar, Hassan, and Alam}]{kmainasi2024llamalens}
Mohamed~Bayan Kmainasi, Ali~Ezzat Shahroor, Maram Hasanain, Sahinur~Rahman Laskar, Naeemul Hassan, and Firoj Alam. 2024.
\newblock {LlamaLens}: Specialized multilingual llm for analyzing news and social media content.
\newblock \emph{arXiv preprint arXiv:2410.15308}.

\bibitem[{Kumar and Nandakumar(2022)}]{kumar2022hate}
Gokul~Karthik Kumar and Karthik Nandakumar. 2022.
\newblock {Hate-CLIPper}: Multimodal hateful meme classification based on cross-modal interaction of clip features.
\newblock In \emph{Proceedings of the Second Workshop on NLP for Positive Impact (NLP4PI)}, pages 171--183.

\bibitem[{Kumari et~al.(2024)Kumari, Jain, and Ekbal}]{kumari2024m3hop}
Gitanjali Kumari, Kirtan Jain, and Asif Ekbal. 2024.
\newblock M3hop-cot: Misogynous meme identification with multimodal multi-hop chain-of-thought.
\newblock \emph{arXiv preprint arXiv:2410.09220}.

\bibitem[{Li et~al.(2022)Li, Chen, Shen, Chen, Zhang, Li, Wang, Qian, Peng, Mao et~al.}]{li2022explanations}
Shiyang Li, Jianshu Chen, Yelong Shen, Zhiyu Chen, Xinlu Zhang, Zekun Li, Hong Wang, Jing Qian, Baolin Peng, Yi~Mao, et~al. 2022.
\newblock Explanations from large language models make small reasoners better.
\newblock \emph{arXiv preprint arXiv:2210.06726}.

\bibitem[{Magister et~al.(2022)Magister, Mallinson, Adamek, Malmi, and Severyn}]{magister2022teaching}
Lucie~Charlotte Magister, Jonathan Mallinson, Jakub Adamek, Eric Malmi, and Aliaksei Severyn. 2022.
\newblock Teaching small language models to reason.
\newblock \emph{arXiv preprint arXiv:2212.08410}.

\bibitem[{Nandi et~al.(2024)Nandi, Sharma, and Chakraborty}]{nandi2024safe}
Palash Nandi, Shivam Sharma, and Tanmoy Chakraborty. 2024.
\newblock {SAFE-MEME}: Structured reasoning framework for robust hate speech detection in memes.
\newblock \emph{arXiv preprint arXiv:2412.20541}.

\bibitem[{O'Neill(2017)}]{o2017overview}
Thomas~A O'Neill. 2017.
\newblock An overview of interrater agreement on likert scales for researchers and practitioners.
\newblock \emph{Frontiers in psychology}, 8:777.

\bibitem[{OpenAI(2023)}]{openai2023gpt4}
OpenAI. 2023.
\newblock \href {https://arxiv.org/abs/2303.08774} {{GPT}-4 technical report}.
\newblock Technical report, OpenAI.

\bibitem[{Piskorski et~al.(2023)Piskorski, Stefanovitch, Nikolaidis, Da~San~Martino, and Nakov}]{piskorski-etal-2023-multilingual}
Jakub Piskorski, Nicolas Stefanovitch, Nikolaos Nikolaidis, Giovanni Da~San~Martino, and Preslav Nakov. 2023.
\newblock \href {https://doi.org/10.18653/v1/2023.acl-long.169} {Multilingual multifaceted understanding of online news in terms of genre, framing, and persuasion techniques}.
\newblock In \emph{Proceedings of the 61st Annual Meeting of the Association for Computational Linguistics (Volume 1: Long Papers)}, pages 3001--3022, Toronto, Canada. Association for Computational Linguistics.

\bibitem[{Plaat et~al.(2024)Plaat, Wong, Verberne, Broekens, van Stein, and Back}]{plaat2024reasoning}
Aske Plaat, Annie Wong, Suzan Verberne, Joost Broekens, Niki van Stein, and Thomas Back. 2024.
\newblock Reasoning with large language models, a survey.
\newblock \emph{arXiv e-prints}, pages arXiv--2407.

\bibitem[{Shah et~al.(2024)Shah, Biswas, Agus, Househ, and Zaghouani}]{shah-etal-2024-mememind}
Uzair Shah, Md.~Rafiul Biswas, Marco Agus, Mowafa Househ, and Wajdi Zaghouani. 2024.
\newblock \href {https://doi.org/10.18653/v1/2024.arabicnlp-1.45} {{M}eme{M}ind at {A}r{AIE}val shared task: Generative augmentation and feature fusion for multimodal propaganda detection in {A}rabic memes through advanced language and vision models}.
\newblock In \emph{Proceedings of The Second Arabic Natural Language Processing Conference}, pages 467--472, Bangkok, Thailand. Association for Computational Linguistics.

\bibitem[{Sharma et~al.(2022)Sharma, Alam, Akhtar, Dimitrov, Da~San~Martino, Firooz, Halevy, Silvestri, Nakov, and Chakraborty}]{ijcai2022p781}
Shivam Sharma, Firoj Alam, Md.~Shad Akhtar, Dimitar Dimitrov, Giovanni Da~San~Martino, Hamed Firooz, Alon Halevy, Fabrizio Silvestri, Preslav Nakov, and Tanmoy Chakraborty. 2022.
\newblock \href {https://doi.org/10.24963/ijcai.2022/781} {Detecting and understanding harmful memes: A survey}.
\newblock In \emph{Proceedings of the Thirty-First International Joint Conference on Artificial Intelligence}, IJCAI~'22, pages 5597--5606, Vienna, Austria. International Joint Conferences on Artificial Intelligence Organization.
\newblock Survey Track.

\bibitem[{Shen et~al.(2022)Shen, Wu, Guo, and Huang}]{shen2022shortest}
Hua Shen, Tongshuang Wu, Wenbo Guo, and Ting-Hao Huang. 2022.
\newblock Are shortest rationales the best explanations for human understanding?
\newblock In \emph{Proceedings of the 60th Annual Meeting of the Association for Computational Linguistics (Volume 2: Short Papers)}, pages 10--19.

\bibitem[{Steiner et~al.(2024)Steiner, Pinto, Tschannen, Keysers, Wang, Bitton, Gritsenko, Minderer, Sherbondy, Long et~al.}]{steiner2024paligemma}
Andreas Steiner, Andr{\'e}~Susano Pinto, Michael Tschannen, Daniel Keysers, Xiao Wang, Yonatan Bitton, Alexey Gritsenko, Matthias Minderer, Anthony Sherbondy, Shangbang Long, et~al. 2024.
\newblock Paligemma 2: A family of versatile vlms for transfer.
\newblock \emph{arXiv preprint arXiv:2412.03555}.

\bibitem[{Sun et~al.(2023)Sun, Li, Li, Wu, Guo, Zhang, and Wang}]{sun_text_2023}
Xiaofei Sun, Xiaoya Li, Jiwei Li, Fei Wu, Shangwei Guo, Tianwei Zhang, and Guoyin Wang. 2023.
\newblock \href {https://doi.org/10.18653/v1/2023.findings-emnlp.603} {Text classification via large language models}.
\newblock In \emph{Findings of the Association for Computational Linguistics: {EMNLP} 2023}, pages 8990--9005. Association for Computational Linguistics.

\bibitem[{Van~de Ven et~al.(2022)Van~de Ven, Tuytelaars, and Tolias}]{van2022three}
Gido~M Van~de Ven, Tinne Tuytelaars, and Andreas~S Tolias. 2022.
\newblock Three types of incremental learning.
\newblock \emph{Nature Machine Intelligence}, 4(12):1185--1197.

\bibitem[{Velioglu and Rose(2020)}]{velioglu2020detecting}
Riza Velioglu and Jewgeni Rose. 2020.
\newblock Detecting hate speech in memes using multimodal deep learning approaches: Prize-winning solution to hateful memes challenge.
\newblock \emph{arXiv preprint arXiv:2012.12975}.

\bibitem[{Wang et~al.(2023)Wang, Hee, Awal, Choo, and Lee}]{wang_evaluating_2023}
Han Wang, Ming~Shan Hee, Md~Rabiul Awal, Kenny Tsu~Wei Choo, and Roy Ka-Wei Lee. 2023.
\newblock Evaluating {GPT}-3 generated explanations for hateful content moderation.
\newblock In \emph{Proceedings of the Thirty-Second International Joint Conference on Artificial Intelligence}, pages 6255--6263.

\bibitem[{Wang et~al.(2024)Wang, Bai, Tan, Wang, Fan, Bai, Chen, Liu, Wang, Ge et~al.}]{wang2024qwen2}
Peng Wang, Shuai Bai, Sinan Tan, Shijie Wang, Zhihao Fan, Jinze Bai, Keqin Chen, Xuejing Liu, Jialin Wang, Wenbin Ge, et~al. 2024.
\newblock Qwen2-vl: Enhancing vision-language model's perception of the world at any resolution.
\newblock \emph{arXiv preprint arXiv:2409.12191}.

\bibitem[{Yang et~al.(2023)Yang, Kim, Kim, Ho, Thorne, and Yun}]{yang2023hare}
Yongjin Yang, Joonkee Kim, Yujin Kim, Namgyu Ho, James Thorne, and Se-Young Yun. 2023.
\newblock {HARE}: Explainable hate speech detection with step-by-step reasoning.
\newblock In \emph{Findings of the Association for Computational Linguistics: EMNLP 2023}, pages 5490--5505.

\bibitem[{Zhang et~al.(2024)Zhang, Huang, Jin, and Lu}]{zhang2024vision}
Jingyi Zhang, Jiaxing Huang, Sheng Jin, and Shijian Lu. 2024.
\newblock Vision-language models for vision tasks: A survey.
\newblock \emph{IEEE Transactions on Pattern Analysis and Machine Intelligence}.

\bibitem[{Zhang et~al.(2023)Zhang, Dong, Li, Zhang, Sun, Wang, Li, Hu, Zhang, Wu et~al.}]{zhang2023instruction}
Shengyu Zhang, Linfeng Dong, Xiaoya Li, Sen Zhang, Xiaofei Sun, Shuhe Wang, Jiwei Li, Runyi Hu, Tianwei Zhang, Fei Wu, et~al. 2023.
\newblock Instruction tuning for large language models: A survey.
\newblock \emph{arXiv preprint arXiv:2308.10792}.

\bibitem[{Zhang et~al.(2020)Zhang, Kishore, Wu, Weinberger, and Artzi}]{zhangbertscore}
Tianyi Zhang, Varsha Kishore, Felix Wu, Kilian~Q Weinberger, and Yoav Artzi. 2020.
\newblock {BERTScore}: Evaluating text generation with {BERT}.
\newblock In \emph{International Conference on Learning Representations}.

\end{thebibliography}

\appendix


\section{Annotation Guideline}
\label{sec:app_annotation_guideline}
You will be shown a meme, a label assigned to it, and an explanation for the assigned label. As an annotator, your task is to carefully examine each meme, label, and explanation. Then assess the quality of the explanation provided for the assigned label.
Follow the steps below to ensure a thorough evaluation:

\paragraph{Analyze the Meme}
\begin{itemize}[noitemsep,topsep=0pt,labelsep=.5em] 
    \item Observe the image and read the accompanying text.
    \item Understand the overall message and the potential implications of the meme.
\end{itemize}
\paragraph{Check the Assigned Label}
\begin{itemize}
    \item Check the given label. The label is the result of annotation done by multiple human annotators.
\end{itemize}

\paragraph{Evaluate the Explanation}
\begin{itemize}[noitemsep,topsep=0pt,labelsep=.5em] 
    \item Read the explanation provided for why the meme has been assigned its label.
    \item Assess the explanation based on the metrics below. Each metric is scored on a Likert scale from 1-5. 
\end{itemize}

\textbf{Kindly note that to evaluate the explanation, you do not have to agree or disagree with the given label.}

\subsection{Metrics}

\subsubsection{Informativeness}
Measures the extent to which the explanation provides relevant and meaningful information for understanding the reasoning behind the label. A highly informative explanation offers detailed insights that directly contribute to the justification, while a low-informative explanation may be vague, incomplete, or lacking key details.

\noindent
\textbf{As an annotator, you are judging if the explanation provides enough information to explain the label assigned to the meme.}

\begin{itemize}[noitemsep,topsep=0pt,labelsep=.5em] 
    \item 1 = Not informative: The explanation lacks relevant details and does not help understand why the meme is labeled as such.
    \item 2 = Slightly informative: The explanation provides minimal information, but key details are missing or unclear.
    \item 3 = Moderately informative: The explanation contains some useful details but lacks depth or supporting reasoning.
    \item 4 = Informative: The explanation is well-detailed, providing a clear and meaningful justification for the label.
    \item 5 = Very informative: The explanation is thorough, insightful, and fully justifies the label with strong supporting details.
\end{itemize}

\subsubsection{Clarity}
Assesses how clearly the explanation conveys its meaning. A clear explanation is well-structured, concise, and easy to understand without requiring additional effort. It should be free from ambiguity, overly complex language, or poor phrasing that might hinder comprehension.

\noindent
\textbf{As an annotator, you are judging the language and structure of the explanation. Spelling mistakes, awkward use of language, and incorrect translations will negatively impact this metric.}

\begin{itemize}[noitemsep,topsep=0pt,labelsep=.5em] 
    \item 1 = Very unclear: The explanation is confusing, vague, or difficult to understand.
    \item 2 = Somewhat unclear: The explanation has some clarity but includes ambiguous or poorly structured statements.
    \item 3 = Neutral: The explanation is somewhat clear but may require effort to fully grasp.
    \item 4 = Clear: The explanation is well-structured and easy to understand with minimal ambiguity.
    \item 5 = Very clear: The explanation is highly readable, precise, and effortlessly understandable.
\end{itemize}

\subsubsection{Plausibility}
Refers to the extent to which an explanation logically supports the assigned label and appears reasonable given the meme's content. A plausible explanation should be coherent, factually consistent, and align with the expected reasoning behind the label. While it does not require absolute correctness, it should not contain obvious contradictions or illogical claims.

\noindent
\textbf{As an annotator, you are judging if the explanation actually supports the label assigned to the meme. For example, if a meme is labeled as Not Propaganda, the explanation given should justify that label.}

\begin{itemize}[noitemsep,topsep=0pt,labelsep=.5em] 
    \item 1 = Not plausible at all: The explanation does not align with the label and seems completely incorrect.
    \item 2 = Weakly plausible: The explanation has some relevance but lacks strong justification or contains logical inconsistencies.
    \item 3 = Moderately plausible: The explanation somewhat supports the label but may be incomplete or partially flawed.
    \item 4 = Plausible: The explanation logically supports the label and is mostly reasonable.
    \item 5 = Highly plausible: The explanation is fully aligned with the label and presents a strong, logical justification.
\end{itemize}

\subsubsection{Faithfulness}
Measures how accurately an explanation reflects the reasoning behind the assigned label. A faithful explanation correctly represents the key factors and logical steps that justify the label, without adding misleading or unrelated details. High faithfulness means the explanation stays true to the actual reasoning used for classification, ensuring reliability and consistency.

\textbf{As an annotator, you are judging how well the explanation reflects the logic behind the label. For example, if the explanation claims an implication of the meme, it should also present the logical reasoning behind it.}

\begin{itemize}[noitemsep,topsep=0pt,labelsep=.5em] 
    \item 1 = Not faithful at all: The explanation is completely unrelated to the given label and does not reflect a valid reasoning process.
    \item 2 = Weakly faithful: Some elements of the explanation are relevant, but much of it is misleading, inconsistent, or lacks proper justification.
    \item 3 = Moderately faithful: The explanation captures parts of the reasoning but includes unrelated, unclear, or unnecessary justifications.
    \item 4 = Faithful: The explanation aligns well with the reasoning behind the label and includes relevant, logical details.
    \item 5 = Highly faithful: The explanation fully and accurately reflects the correct reasoning, without any misleading or irrelevant information.
\end{itemize}

\section{Annotation Platform}
\label{sec:app_annotation_platform}

\begin{figure*}[htb!]
    \centering
    \includegraphics[scale=0.30]{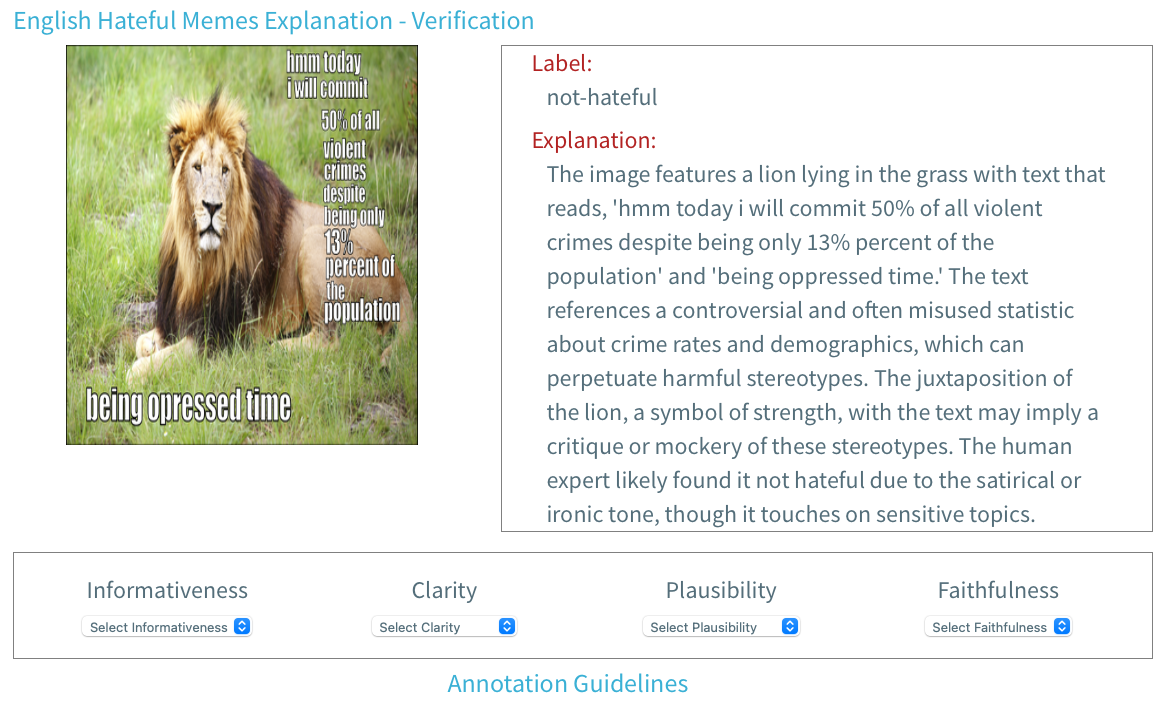}
    \caption{A screenshot of the annotation platform for the explanation evaluation of hateful meme.}
    \label{fig:hateful_meme_annotation_interface}
\end{figure*}

In Figure \ref{fig:hateful_meme_annotation_interface}, we present the screenshot of the interface designed for the explanation evaluation of hateful meme, which consisted of an image, respective label, and explanation for the label, annotation guidelines, and four different evaluation metrics. We used 5-point Likert scale for each evaluation metric. Annotators select one of the Likert scale value following the annotation guideline for each metric and submit.

\section{Prompt for Explanation Generation}
\label{sec:app_prompt}
In Listings \ref{lst:prompt_explanation_generation_armeme_ar_expl} and \ref{lst:prompt_explanation_generation}, we provide the prompts used to generate explanations for ArMeme and Hateful Meme. The prompt in Listing \ref{lst:prompt_explanation_generation_armeme_ar_expl} is specifically for generating Arabic explanations for ArMeme. To generate English explanations, the same prompt was used, except it was adapted for English. 

\begin{lstlisting}[caption={Prompt for generating Arabic explanation for ArMeme dataset. The place-holders refers to label, and number of words, respectively.},label={lst:prompt_explanation_generation_armeme_ar_expl}]
You are a Propaganda Image Detection Expert. A human expert has already classified the image as {class_label}. 
Do not change or re-identify the classified label of the image.
First, analyze the image's visual elements (objects, symbols, color usage, composition) 
and provide a concise description. Next, read and paraphrase any text in the image-especially if 
it is in non-Arabic-so that a Arabic speaker could understand its literal or intended meaning. 
Explain how that text, in conjunction with the visual elements, supports or illustrates the known 
propaganda technique. Think carefully about emotional appeals, humor, cultural references, manipulative 
language, or other rhetorical strategies.
Finally, provide a thoughtful explanation in Arabic (up to {explanation_length} words) of why these elements led the human to classify 
the image as propaganda. Be specific about how the text and visuals interact, referencing the 
Arabic context or cultural nuances if relevant. Your answer must be valid JSON with a single field:
{{
"explanation": "Your explanation here (up to {explanation_length} words) in Arabic."
}}
\end{lstlisting}

\begin{lstlisting}[caption={Prompt for generating explanation. The place-holders refers to label, and number of words, respectively.},label={lst:prompt_explanation_generation}]
You are a Hateful Image Detection Expert. A human expert has already classified the image as {}. Do not change or re-identify the classified label of the image.
First, analyze the image's visual elements (objects, symbols, color usage, composition) and provide a concise description. Next, read and paraphrase any text in the image so anyone speaking English can understand its literal or intended meaning. 
Explain how that text, in conjunction with the visual elements, supports or illustrates the known hateful content. 
Finally, provide a thoughtful explanation in English (up to {} words) of why these elements led the human to classify the image as hateful. Be specific about how the text and visuals interact, referencing the context or cultural nuances if relevant. Your answer must be valid JSON with a single field:
{{
"explanation": "Your explanation here (up to {} words) in English."
}}
\end{lstlisting}


\section{Error Analysis}
\label{ssec:app_error_analysis}

\subsection{Correct vs. Incorrect Label Prediction}
\begin{figure*}[tbh!]
    \centering
    \begin{subfigure}[b]{0.85\textwidth}
        \centering
        \includegraphics[width=\textwidth]{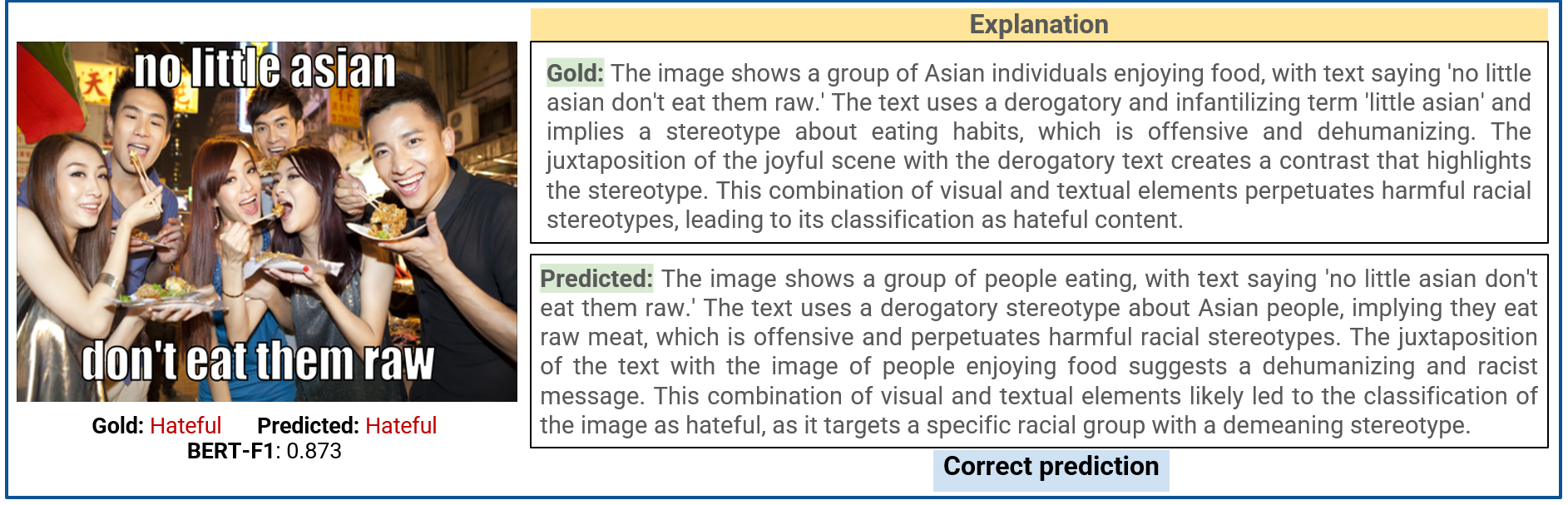}
        \caption{Correct prediction.}
        \label{fig:correct_prediction}
    \end{subfigure}
    \vspace{0.5cm}
    \begin{subfigure}[b]{0.85\textwidth}
        \centering
        \includegraphics[width=\textwidth]{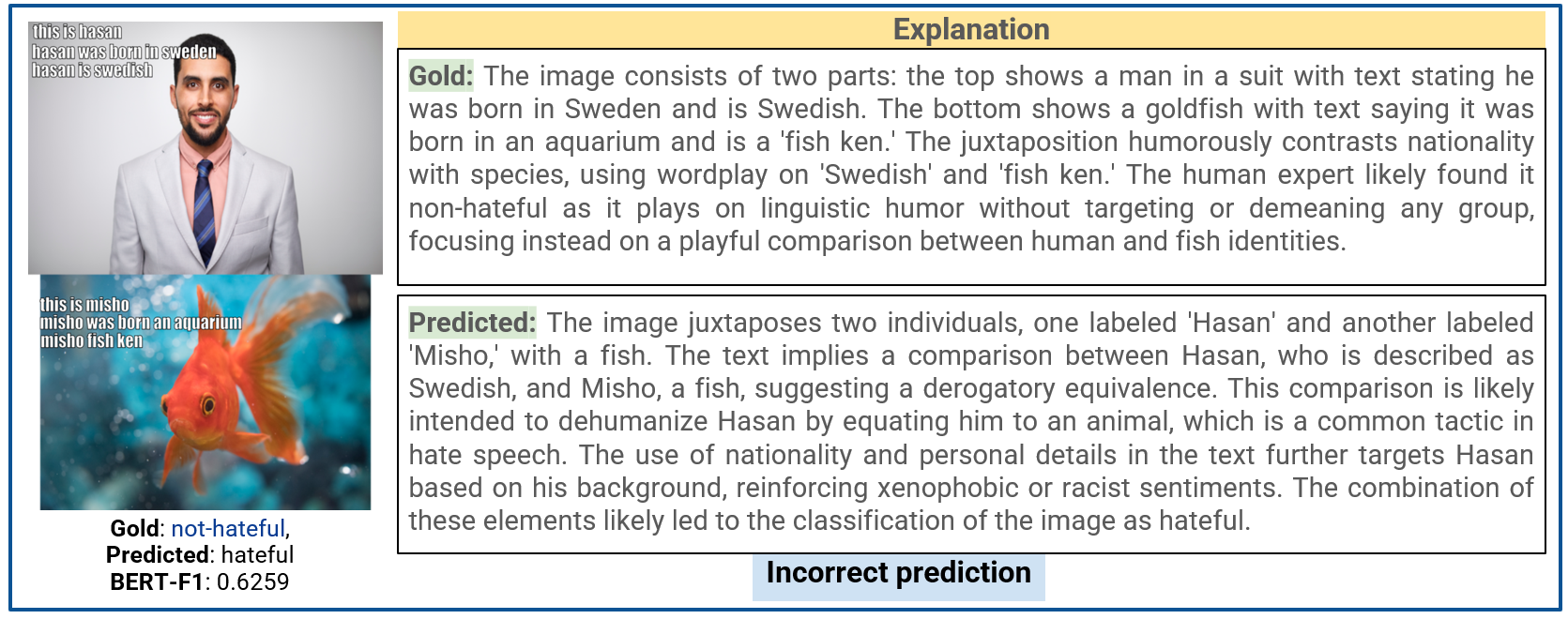}
        \caption{Incorrect prediction.}
        \label{fig:incorrect_prediction}
    \end{subfigure}
    \caption{Example of correct and incorrect label prediction with explanation.}
    \label{fig:correct_incorrect_prediction}
\end{figure*}

In Figure \ref{fig:correct_incorrect_prediction}, we present examples from the Hateful Memes dataset, showcasing cases where the model made both correct and incorrect predictions.

In Figure \textbf{\ref{fig:correct_prediction}}, the Gold explanation describes the image as reinforcing a harmful racial stereotype by juxtaposing a joyful scene of Asian individuals eating with offensive text. The Predicted explanation correctly identifies the derogatory language and its racist implications, aligning with the gold annotation. The model’s BERT-F1 score of 0.873 shows the high confidence in associating textual and visual elements to detect hate speech effectively.

In Figure \textbf{\ref{fig:correct_incorrect_prediction}}, the Gold explanation interprets the image as a humorous juxtaposition, using wordplay between nationality and species without targeting any group. However, the Predicted explanation classifies it as hateful. This missclassification suggests that the model struggled to distinguish linguistic humor from implicit hate speech, as reflected in its BERT-F1 score of 0.6259. This highlights the challenge of detecting context-dependent content, where intent and interpretation play a crucial role in classification.

\subsection{Effect of Single vs. Multistage Training}
\begin{figure*}[htb!]
    \centering
    \includegraphics[width=0.9\textwidth]{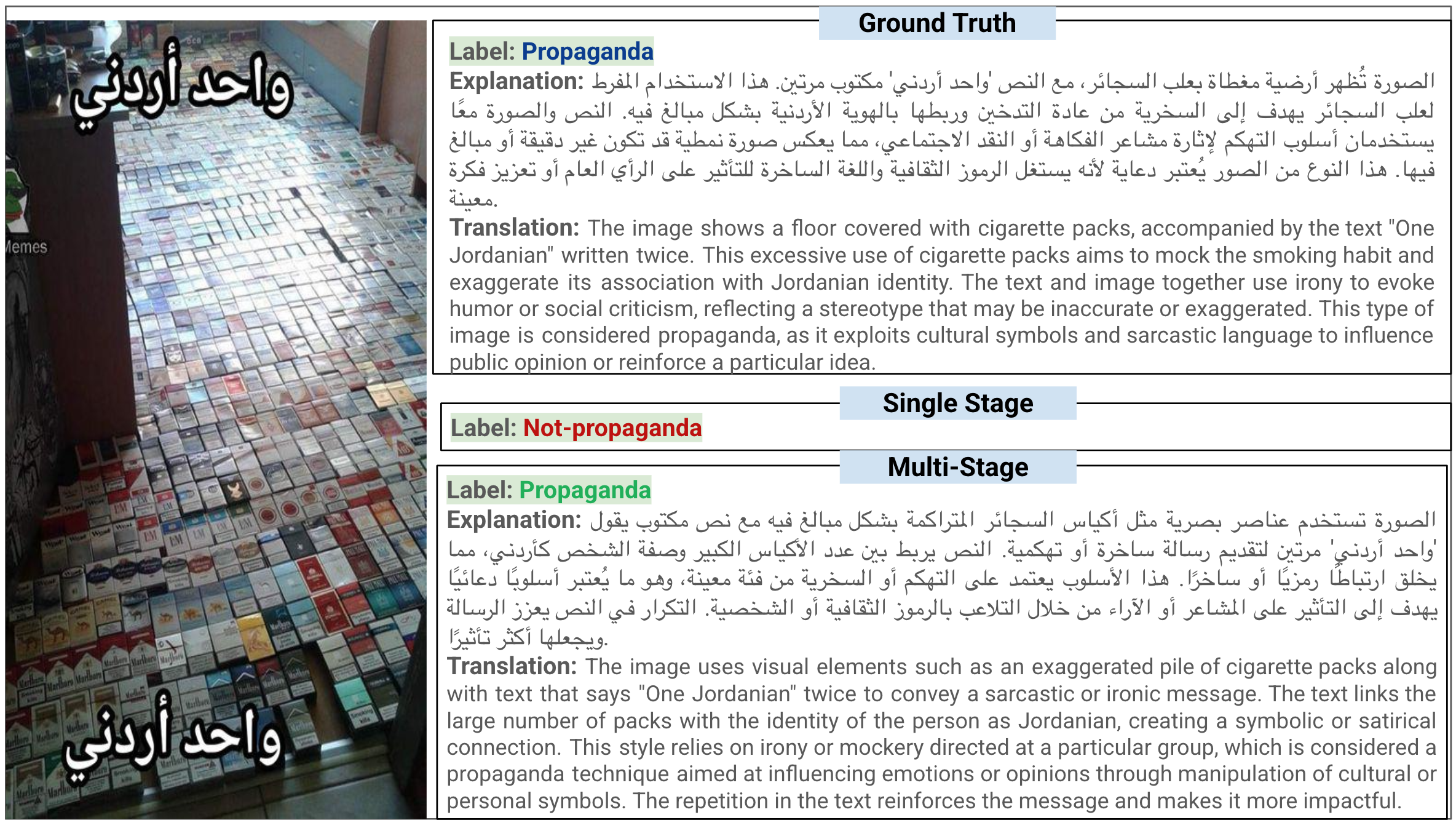}
    \caption{Single- vs. Multi-Stage prediction for \textit{ArMeme}.}
    \label{fig:error_analysis1}
\end{figure*}

\begin{figure*}[]
    \centering
    \includegraphics[width=0.9\textwidth]{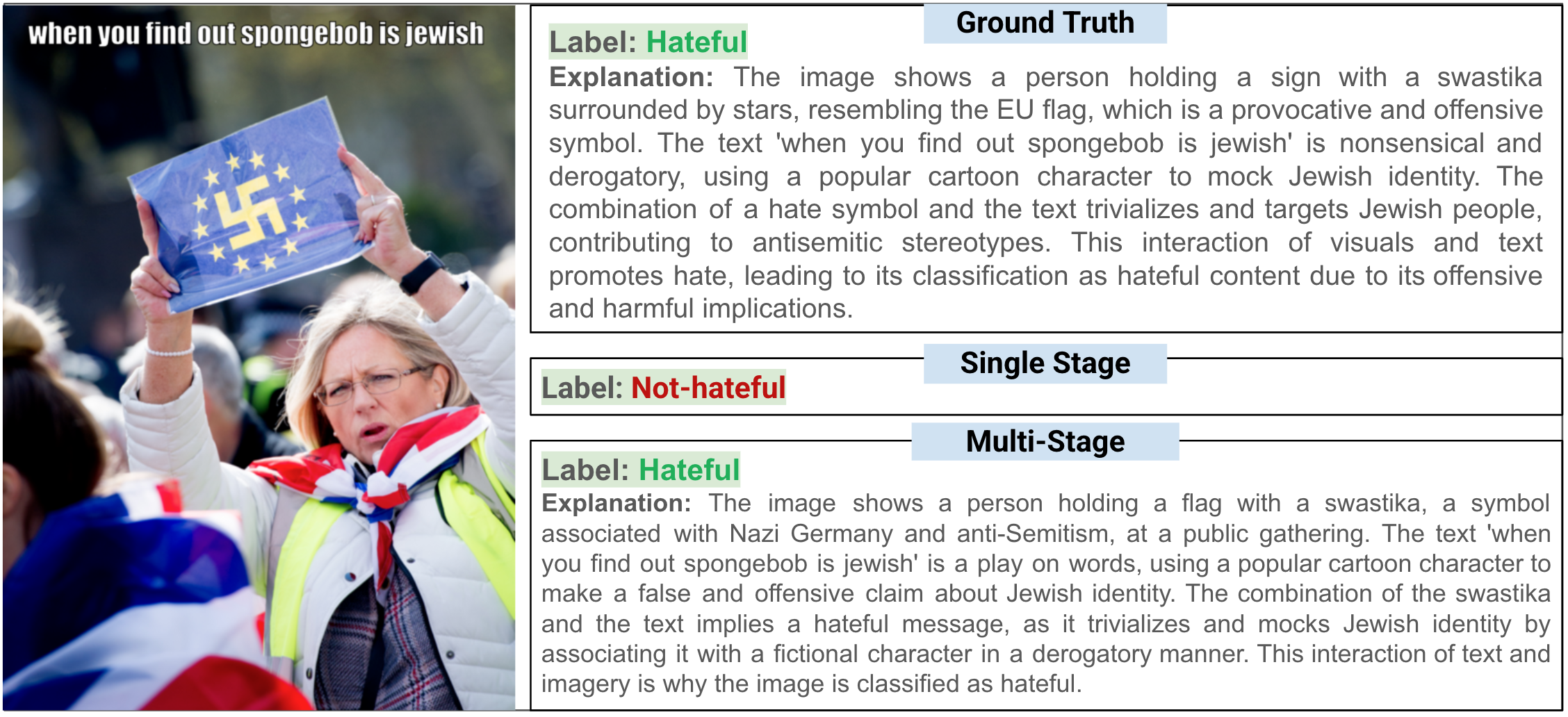}
    \caption{Single- vs. Multi-Stage prediction for \textit{hateful memes}.}
    \label{fig:error_analysis2}
\end{figure*}

In Figure \textbf{\ref{fig:error_analysis1}} we notice that multi-stage model correctly predicts the label and generates a faithful explanation, identifying how the combination of the image and text reinforces gender stereotypes. In contrast, the single-stage model label the meme with incorrect label -- not-propaganda. Figure \textbf{\ref{fig:error_analysis2}} shows another example from the hateful memes dataset. The multi-stage model correctly labels the image as hateful and grounds its explanation in the offensive combination of the \textit{swastika} and \textit{the mocking caption}. Whereas the single-stage model misclassifies the image as not-hateful. This highlights the effectiveness of multistage optimization in aligning predictions with grounded reasoning.



\section{Data Release}
\label{apndix:release}
The \memex{} dataset\footnote{\url{anonymous.com}} will be released under the CC BY-NC-SA 4.0 -- Creative Commons Attribution 4.0 International License: \url{https://creativecommons.org/licenses/by-nc-sa/4.0/}.


\end{document}